\documentclass{article}

\PassOptionsToPackage{numbers,compress}{natbib}
\usepackage[preprint]{neurips_2026}

\usepackage[utf8]{inputenc} 
\usepackage[T1]{fontenc}    
\usepackage{hyperref}       
\usepackage{url}            
\usepackage{booktabs}       
\usepackage{amsfonts}       
\usepackage{nicefrac}       
\usepackage{microtype}      
\usepackage{xcolor}         

\usepackage{graphicx}
\hypersetup{
    colorlinks=true,
    linkcolor=blue,
    citecolor=blue,
    urlcolor=blue
}
\usepackage{colortbl}
\usepackage{booktabs}
\usepackage{makecell}
\usepackage{pgf}
\usepackage{colortbl}
\usepackage{booktabs}
\usepackage[dvipsnames]{xcolor}
\usepackage{makecell}
\usepackage{pgf}
\usepackage{caption, subcaption}
\usepackage{soul}
\usepackage{xspace}
\usepackage[breakable,most]{tcolorbox}
\newcommand{\sftmodel}{\textsc{SFTuser}\xspace}
\newcommand{\sftmodelshort}{\textsc{SFT}\xspace}
\newcommand{\sftmodeltwo}{\textsc{SFTuser2}\xspace}
\newcommand{\roleplay}{\textsc{RPuser}\xspace}
\newcommand{\roleplayshort}{\textsc{RP}{\scriptsize 3}\xspace}
\newcommand{\roleplayone}{\textsc{RPuser}{\scriptsize 1}\xspace}
\newcommand{\roleplaytwo}{\textsc{RPuser}{\scriptsize 2}\xspace}
\newcommand{\roleplaythree}{\textsc{RPuser}{\scriptsize 3}\xspace}
\newcommand{\nouser}{\textsc{Static}\xspace}

\usepackage{enumitem}
\usepackage{array}
\usepackage{amsmath} 
\usepackage{multirow}
\usepackage{float}
\usepackage{tabularx}
\usepackage{booktabs}

\newlength{\myneg}
\newlength{\figneg}
\title{
Quantifying the Utility of User Simulators \\ for Building Collaborative LLM Assistants
}

\author{%
  Joseph Suh\thanks{\texttt{josephsuh@berkeley.edu}, \texttt{serinac@berkeley.edu}} \\
  UC Berkeley \\
  \And
  Ayush Raj \\
  UC Berkeley \\
  \And
  Minwoo Kang \\
  UC Berkeley \\
  \And
  Serina Chang\footnotemark[1] \\
  UC Berkeley \\
}

\begin{document}

\maketitle

\begin{abstract}
\label{sec:abstract}

User simulators are increasingly leveraged to build interactive AI assistants,
yet how to measure the quality of these simulators remains an open question.
In this work, we show how simulator quality can be quantified in terms of its downstream utility:
how an LLM assistant trained with this user simulator 
performs \emph{in the wild} when interacting with real humans.
In a controlled experiment where only the user simulator varies,
we train LLM assistants via reinforcement learning against a spectrum of simulators,
from an LLM prompted to role-play a user to one fine-tuned on human utterances from WildChat.
As evaluation, we measure pairwise win rates in a user study with 283 participants and on WildBench, a benchmark derived from real human--AI conversations.
Training against the role-playing LLM yields an assistant statistically indistinguishable from the initial assistant in our user study (51\% win rate),
whereas training against the fine-tuned simulator yields significant gains (58\% over the initial and 57\% over the one trained against role-playing).
Closer inspection reveals three further patterns:
methods for making role-playing LLMs more realistic (e.g., persona conditioning) improve trained assistants but do not close the gap to the fine-tuned simulator;
scaling the simulator's model size benefits the fine-tuned simulator but yields no gain for role-playing ones; and
assistants trained against role-playing simulators fail to generalize when paired with other simulators at test time,
while the one trained against fine-tuned simulator does.
Together, these results argue for grounding user simulators in real human behavior
and measuring their quality by their downstream effect on real users.
\footnote{ We release our code, trained LLM assistant models, learned user simulator models, and simulated conversation trajectories at \href{https://github.com/schang-lab/utility-of-user-simulators}{https://github.com/schang-lab/utility-of-user-simulators}.}
\end{abstract}
\section{Introduction}
\label{sec:intro}

\begin{figure}[t]
    \centering
    \includegraphics[width=1.00\linewidth]{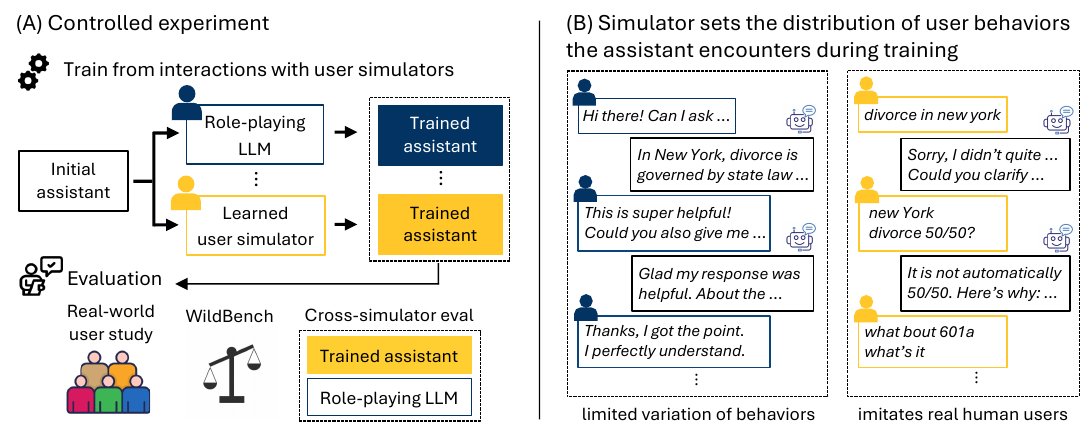}
    \vspace{-20pt}
    \caption{
    \textbf{(A)}
    For each user simulator, we train an assistant via RL from its interactions with the simulator, then evaluate trained assistants in three ways:
    real-world user study, real-world task benchmark (WildBench), and cross-simulator evaluation.
    \textbf{(B)}
    Two simulated conversation trajectories: the user simulator determines the distribution of user behaviors the assistant sees during training.
    }
    \vspace{\figneg}
    \label{fig:figure_prelim}
\end{figure}

A good AI assistant should interact effectively with human users across multiple turns of conversation \citep{laban2025llms,laidlaw2025assistancezero}.
Training and evaluating such assistants require multi-turn human–AI interactions,
but access to human users is often limited \citep{chiang2024chatbot}.
Recent works have therefore explored building LLM-based \textit{user simulators},
where the simulator produces the user's turns in a conversation with the assistant
\citep{ni2026survey}.
Simulation offers data collection that is scalable, cost-effective, and free from safety risks of deploying undertrained models to interact with humans \citep{peng2018sim}.

Yet user simulators vary in how closely they approximate human users.
The most common approach
is an instruction-tuned LLM prompted to role-play as a human user~\citep{park2023generative, yao2024tau, li2024iqa, luo2024duetsim, gao2024regressing, shani2024multi, zhang2024modeling, kong2024platolm, wu2025collabllm, sun2025training, dinucu2025problem, qian2025userrl}.
This approach is simple and does not require human data, but produces less realistic user turns such as overly cooperative and well-structured utterances \citep{yoon2024evaluating, wang2025human}.
Given this realism gap, recent work has sought to develop more realistic simulators, from initializing simulated conversations with the first turn from a human user \citep{ivey2024robotic}
to crafting diverse personas \citep{ge2024scaling,dou-etal-2025-simulatorarena, nvidia/Nemotron-Personas-Korea} to fine-tuning on human data \citep{wang2025know,naous2025flipping,chang2025chatbench,wu2026humanlm}.
In these studies, various metrics have been proposed to quantify simulator quality via similarity between simulated and human utterances
(e.g., perplexity or overlap in linguistic features)
\citep{zhou2026mind}.
However, the way simulator quality should be measured remains contested.

We propose measuring simulator quality by what simulators are ultimately used for.
One increasingly important motivation for user simulators is training assistants that perform well with human users
\citep{li2024iqa, luo2024duetsim, gao2024regressing, shani2024multi, zhang2024modeling, kong2024platolm, wu2025collabllm, sun2025training, dinucu2025problem, qian2025userrl, levin2000stochastic, xu2016policy, takanobu2019guided, keizer2023adversarial, carroll2019utility}.
Hence a natural way to define simulator quality is in terms of its downstream utility:
\textbf{how an LLM assistant trained with the user simulator performs when tested with real humans}.
We operationalize this with a controlled experiment (\autoref{fig:figure_prelim}a).
Concretely, we train LLM assistants via reinforcement learning (RL) from interactions with user simulators,
then evaluate them in interactions with human users.
We compare several simulators spanning a spectrum of design choices,
from a role-playing LLM (\roleplay) to a learned user model (\sftmodel) trained via supervised fine-tuning on user utterances from WildChat \citep{naous2025flipping, zhao2024wildchat}, a corpus of human--LLM conversations.
For each simulator, we train one assistant via multi-turn RL,
with rewards from LLM judges that score its interactions with the simulator.
We evaluate the trained assistants in a user study,
where human participants interact with assistants and provide pairwise preferences,
and on WildBench \citep{lin2024wildbench},
a held-out set of difficult real-world tasks from WildChat users.

We find that the assistant trained with the role-playing LLM suffers from a severe train-test mismatch (Section~\ref{sec:user_study}).
Although the assistant achieves higher rewards against its train-time partner (\roleplay),
these gains fail to transfer to human users,
yielding only a 50.6$\pm$4.2\% win rate against the initial untrained assistant in our user study.
By contrast, the assistant trained with the learned user model (i.e., \sftmodel-trained assistant) generalizes to human users, winning 58.1$\pm$3.8\% against the initial and 57.3$\pm$3.8\% against the \roleplay-trained assistant.
We attribute the \roleplay-trained assistant's failure to a \emph{distribution shift} between training and deployment,
induced by the unrealistic behavior of the role-playing LLM (Figure~\ref{fig:figure_prelim}b).
The gap is most pronounced on open-ended tasks such as creative writing, where multi-turn drift from realistic user behavior is most damaging.

A more nuanced picture emerges on closer inspection.
First,
methods to make role-playing LLMs more realistic, such as persona conditioning,
improve the trained assistant's performance but still underperform the \sftmodel-trained counterpart (Section~\ref{sec:experiments:wildbench}).
Second,
scaling the model size of role-playing LLMs does not yield performance gains for the trained assistant,
but scaling the \sftmodel does,
suggesting that the two simulator types differ fundamentally in how they benefit from scale (Section~\ref{sec:experiments:user_sim_metrics}).
Finally,
evaluating assistants against simulators they were not trained with reproduces the generalization gap observed in the user study.
Assistants trained with role-playing LLMs drop sharply when paired with \sftmodel,
whereas the assistant trained with \sftmodel generalizes to role-playing LLMs it never encountered during training.
This suggests that evaluating across simulators can serve as a practical proxy for generalization to human users (Section~\ref{sec:experiments:cross_environment}).

As LLM assistants are increasingly trained from simulated interactions,
the choice of user simulator directly shapes the assistant that human users ultimately interact with.
Our work makes the case for grounding simulators in real human behavior rather than role-played approximations,
and for measuring their utility by what ultimately matters---how the trained assistants perform with humans.
\section{Training Assistants from Simulated Interactions}
\label{sec:prelim}
\vspace{\myneg}

We study open-domain conversational assistants as a prominent instance of human--AI interaction.
We use WildChat-1M~\citep{zhao2024wildchat} for the assistant training pipeline,
and additionally use WildBench~\citep{lin2024wildbench} and WildChat-4.8M~\citep{deng2024wildvis} for evaluation.
Although our experiments focus on open-domain conversation,
the framework applies to any multi-turn human--AI interaction.

\textbf{Preliminaries.}
A simulated conversation is a sequence of alternating turns:
an utterance generated by the user simulator followed by a response from the LLM assistant.
A length-$n$ conversation consists of $2n$ turns.
The conversation ends when the user simulator emits a special end-conversation token or when a maximum turn limit is reached;
we set this limit to $n=5$, which covers 91\% of conversations in WildChat.
Let $o_t = \{u_1, a_1, \dots, u_t\}$ for $1 \leq t \leq n$ denote the
conversation trajectory up to the $t$-th user utterance $u_t$. From an RL
perspective, $o_t$ is the observation visible to the agent (the LLM assistant)
at time $t$, and the $t$-th assistant response $a_t$ is the action taken by the agent
given $o_t$.

Each conversation is associated with a high-level user intent $z$ describing what the user aims to achieve from the conversation (e.g., rewriting a text concisely) \citep{pietquin2006probabilistic}.
Following prior work~\citep{naous2025flipping, wu2026humanlm, lin2024wildbench},
we extract intents from WildChat-1M conversations by prompting Qwen3-32B to summarize the user's high-level goal from the conversation history (Appendix~\ref{sec:appendix:wildchat}).
While the user simulator is conditioned on $z$ to simulate utterances, the assistant observes only $o_t$ to generate responses.
This partial observability \citep{laidlaw2025assistancezero, fern2014decision} is far from a toy abstraction;
real users' intents are typically underspecified at the outset
and only partially expressible in any single utterance
\citep{herlihy2024overcoming, yang2025prompts}.

\textbf{User as the transition dynamics.}
Given $o_t$ and $a_t$, the distribution over $o_{t+1}$ is governed by how the user (either human or simulator) reacts to the assistant response, making the user the environment's transition function.
This view reveals a source of the sim-to-real gap when training assistants with user simulators \citep{carroll2019utility}.
Role-playing LLMs tend to produce
predictable utterances~\citep{yoon2024evaluating, zhou2026mind}, yielding a
nearly deterministic transition function. Human users, by contrast, are highly
stochastic transition functions: they rephrase confusingly, skip steps, and express
frustration~\citep{piantadosi2012communicative, naous2025flipping}. An assistant
trained against a role-playing LLM therefore risks overfitting to a
narrow region of the true human--LLM conversation state space and failing to generalize
when paired with real humans \citep{tobin2017domain, peng2018sim, cobbe2019quantifying}.

\textbf{Training the assistant via RL.}
We generate multi-turn conversation rollouts between the simulator and the assistant.
At the end of each rollout, the assistant receives a trajectory-level
reward from LLM judges that score response quality on a Likert-scale rubric~\citep{wu2025collabllm, gunjal2025rubrics,
kim2024prometheus, liu2023g}; full rubric details appear in
\autoref{appendix:sec:rubrics}.
We note that the judges see the high-level user intent $z$ alongside the full trajectory,
while the assistant policy never observes $z$ and acts under partial observability.
This information asymmetry lets the judge assess whether the assistant has addressed the user's underlying goal~\citep{pinto2017asymmetric, lowe2017multi}.

As an alternative that avoids user simulation, the assistant can be
trained via single-turn RL
on static offline human--LLM conversations~\citep{bai2022constitutional, ouyang2022training, bai2022training}.
We present its experimental results in Appendix~\ref{appendix:sec:additional_results}.

\section{User Simulators}
\label{sec:user_simulators}
\vspace{\myneg}

We describe user simulators that can demonstrate the utility of user simulators for training assistants.

\vspace{\myneg}
\subsection{Role-Playing LLM}
\label{sec:user_simulators:role_playing}
\vspace{\myneg}

The dominant type of user simulator that has been used to train LLM assistants is an LLM prompted to role-play as a user \citep{ni2026survey}.
We consider three variants of role-playing LLMs, each hypothesized to approximate human utterance more faithfully than the last.

\textbf{\roleplayone:}
An LLM simulates the user from the first turn onward, guided by an instruction describing the user role and an intent $z$~\citep{li2024iqa, luo2024duetsim, kong2024platolm, wu2025collabllm, dinucu2025problem}.
The instruction includes details such as ``You are role-playing as a human USER'' and ``As a user, avoid being too detailed in your responses'';
for a full prompt, please refer to~Appendix \ref{appendix:sec:baseline_user_model:fully_synthetic}.
We note that \roleplayone is the most widely adopted approach for role-playing user simulators, as it is simple and requires no human data~\citep{yao2024tau, li2024iqa, luo2024duetsim, gao2024regressing, shani2024multi, zhang2024modeling, kong2024platolm, wu2025collabllm, sun2025training, dinucu2025problem, qian2025userrl}.

\textbf{\roleplaytwo:}
The opening user query is drawn from real first-turn user utterances in WildChat, but all subsequent user utterances are simulated by the role-playing LLM \citep{ivey2024robotic}.
\roleplaytwo has a representative initial utterance distribution but relies on the role-playing LLM for later turns (Appendix~\ref{appendix:sec:baseline_user_model:human_seeded}).

\textbf{\roleplaythree:}
Building on prior work that uses persona prompts to simulate diverse populations~\citep{ge2024scaling, hu2024quantifying, castricato2025persona}
and taxonomies of conversational behavior~\citep{shim2025non},
we inject persona descriptions into the instruction.
For each simulated conversation, we randomly sample a description (e.g., ``Be Impatient: Expect fast, to-the-point answers'') from a persona pool (Appendix~\ref{appendix:sec:baseline_user_model:persona}).

\vspace{\myneg}
\subsection{Learned User Simulator With Supervised Fine-Tuning}
\label{sec:user_simulators:sft}
\vspace{\myneg}

Following UserLM~\citep{naous2025flipping} and user simulators for conversational recommender system~\citep{meshi2026convapparel}, we fine-tune a language model on the user utterances of WildChat-1M.
It is the simplest recipe for building user simulators when human--AI interaction data is available \citep{bain1995framework,levin2000stochastic};
we leave exploration of other methods (e.g., IRL, GAIL) \citep{ng2000algorithms, ho2016generative} for future work.
For implementation details, including preprocessing steps such as language filtering and deduplication, please refer to \autoref{appendix:sec:behavioral_cloning}.

This model (which we call \sftmodel) is the learned user simulator available during assistant training.
However, WildChat-1M only includes conversations with specific LLM assistants (OpenAI's GPT-3.5-Turbo and GPT-4) and over a limited time period, from April 2023 to May 2024.
To test generalization of trained assistants to users interacting with other assistants or in other time periods,
we prepare a second model \sftmodeltwo,
trained on WildChat-4.8M whose data collection period and interacting LLM assistants are not overlapping with those of WildChat-1M.
\sftmodeltwo is available only at test time to evaluate the trained assistants' generalizability to a new user simulator.
\section{Experiments}
\label{sec:experiments}
\vspace{\myneg} 

In this section, we describe the experimental setup (Section~\ref{sec:experiments:setup}) and results of our main experiments:
\vspace{\myneg}
\begin{enumerate}[leftmargin=2em, itemsep=2pt, parsep=0pt]
    \item Evaluations on WildBench, to evaluate how assistants trained with different user simulators perform on tasks from real human users (Section~\ref{sec:experiments:wildbench}),
    \item Examining how scaling the size of the LLM underlying the user simulator — both fine-tuned and role-playing — affects the performance of the trained assistant (Section~\ref{sec:experiments:user_sim_metrics}),
    \item Cross-simulator evaluation, where we evaluate how well each assistant trained with one user simulator generalizes to another user simulator at test time (Section~\ref{sec:experiments:cross_environment}).
\end{enumerate}
\vspace{\myneg}
We focus on automated evaluations in this section,
which enable us to run multiple reproducible experiments,
but provide results from our user study with human participants in Section~\ref{sec:user_study}.

\vspace{\myneg}
\subsection{Experimental Setup}
\label{sec:experiments:setup}
\vspace{\myneg}

We conduct a controlled experiment in which trained assistants differ \emph{only} in the user simulator they encountered during training.
The rubric, the judge model, and the initial assistant are held constant across all conditions.
This design isolates the effect of the user simulator from confounding factors such as rubric variation, LLM judges, sampling parameters, and RL algorithms.

\textbf{User simulator.}
All user simulators are derived from the same LLM, Qwen2.5-14B-Instruct \citep{bai2023qwen}.
Specifically, we prepare \sftmodel (Section~\ref{sec:user_simulators:sft}) by fine-tuning it on WildChat-1M with an 89/5/6\% train/validation/test split.
The three role-playing \roleplayone--\roleplaythree (Section~\ref{sec:user_simulators:role_playing}) also use it.

\textbf{LLM assistant.}
We initialize the LLM assistant from Qwen2.5-\{1.5,3\}B-Instruct.
For each initial LLM, we obtain multiple assistants by training with each user simulator.
All comparisons in the subsequent sections are made among assistants derived from Qwen2.5-3B-Instruct; the result with Qwen2.5-1.5B-Instruct is presented in \autoref{appendix:sec:additional_results}.
 
\textbf{Judge.}
During train, we use two open-source LLMs as judges:
Qwen3-32B \citep{yang2025qwen3technicalreport} and Mistral-Small-3.1-24B-Instruct-2503 \citep{jiang2023mistral7b}.
The reward for each simulated conversation trajectory is the average of the two judges' scores, each an integer in 0--10.
Since an assistant optimized against a given LLM judge can overfit to its biases (\autoref{appendix:sec:extended_related_work}),
evaluation on WildBench (Section~\ref{sec:experiments:wildbench}) uses a disjoint set of judges:
majority voting with gpt-5-mini, gemini-2.5-flash, and claude-haiku-4-5-20251001.
 
\textbf{Algorithm.}
We use GRPO \citep{shao2024deepseekmath}.
The trajectory-level reward is provided by the LLM judges described above, and the assistant policy is trained until convergence of the average reward on the validation set.
RL details are in \autoref{appendix:sec:rl_detail},
and the example training curves are in \autoref{fig:figure_rl_training_curve_userlm} and \ref{fig:figure_rl_training_curve_advanced}.

\vspace{\myneg}
\subsection{Evaluating Performance on Real-World Tasks with WildBench}
\label{sec:experiments:wildbench}
\vspace{\myneg}

How do LLM assistants trained with each user simulator perform on real-world tasks?
In this section, we operationalize `real-world' via WildBench, a benchmark constructed with a held-out set of real human--LLM conversations from WildChat.

\textbf{WildBench.}
WildBench-v2 \citep{lin2024wildbench} contains 1,024 human--LLM conversations held out from WildChat release.
For a length-$n$ conversation,
WildBench provides the first $2n{-}1$ turns as context and prompts the assistant to generate the $n$-th response.
Each conversation is paired with a tailored checklist of 11.4 items on average (e.g., ``Does the AI's response acknowledge the user dissatisfaction with the TV shows except for the Martian Manhunter one?'') used to judge response quality.
We report two metrics:
the checklist satisfaction rate of each assistant and,
for pairwise comparisons, the fraction of conversations in which one assistant satisfies more checklist items than the other assistant.
WildBench exhibits a strong correlation ($r^2=0.94$) with human-voted Elo ratings on Chatbot Arena \citep{zheng2023judging}, making it a useful proxy for assistant performance in real-world settings.

\definecolor{heatlo}{HTML}{F5C4B3}
\definecolor{heathi}{HTML}{9FE1CB}
\definecolor{diaggray}{HTML}{EEEEEE}
\definecolor{cigray}{HTML}{777777}

\newcommand{\pcell}[2]{%
    \pgfmathtruncatemacro{\rawpct}{min(100, max(0, ((#1 - 30) / 40) * 100))}%
    \edef\heatspec{heathi!\rawpct!heatlo}%
    \expandafter\cellcolor\expandafter{\heatspec}%
    {\small #1{\scriptsize\,$\pm$\,#2}}%
}
\newcommand{\pcellbf}[2]{%
    \pgfmathtruncatemacro{\rawpct}{min(100, max(0, ((#1 - 30) / 40) * 100))}%
    \edef\heatspec{heathi!\rawpct!heatlo}%
    \expandafter\cellcolor\expandafter{\heatspec}%
    {\small\textbf{#1}{\scriptsize\,$\pm$\,#2}}%
}
\newcommand{\diagcell}{\cellcolor{diaggray}\textcolor{cigray}{---}}

\newcommand{\rrpsynth}{\textsc{RP}\textsubscript{1}}
\newcommand{\rrpseed}{\textsc{RP}\textsubscript{2}}
\newcommand{\rrppers}{\textsc{RP}\textsubscript{3}}
\newcommand{\rot}[1]{\makecell{#1}}

\begin{table}[t]
    \centering
    \small
    \setlength{\tabcolsep}{2.0pt}
    \renewcommand{\arraystretch}{1.3}
    \caption{
        Tie-accounted pairwise win rate matrix and checklist satisfaction rate on WildBench-v2.
        Entry $(i, j)$ gives the win rate of the assistant trained with $j$ against the one trained with $i$.
        For win rates, values following $\pm$ denote the half-width of the 95\% Wald confidence interval;
        for satisfaction rates, they denote the standard error of the mean.
        In each row, the best performance is bold-faced.
    }
    \begin{tabular}{r *{5}{w{c}{4.5em}}}
        \toprule
        \textit{Win} [\%]
        & \rot{Initial} & \rot{\roleplayone} & \rot{\roleplaytwo} & \rot{\roleplaythree} & \rot{\sftmodel} \\
        \midrule
        Initial
            & \diagcell
            & \pcell{55.4}{2.7}
            & \pcell{59.9}{2.7}
            & \pcell{61.3}{2.7}
            & \pcellbf{68.4}{2.6} \\
        \roleplayone
            & \pcell{44.6}{2.7}
            & \diagcell
            & \pcell{55.2}{2.8}
            & \pcell{56.6}{2.8}
            & \pcellbf{63.9}{2.7} \\
        \roleplaytwo
            & \pcell{40.1}{2.7}
            & \pcell{44.8}{2.8}
            & \diagcell
            & \pcell{51.0}{2.8}
            & \pcellbf{60.0}{2.7} \\
        \roleplaythree
            & \pcell{38.7}{2.7}
            & \pcell{43.4}{2.8}
            & \pcell{49.0}{2.8}
            & \diagcell
            & \pcellbf{58.5}{2.8} \\
        \sftmodel
            & \pcell{31.6}{2.6}
            & \pcell{36.1}{2.7}
            & \pcell{40.0}{2.7}
            & \pcell{41.5}{2.8}
            & \diagcell \\
        \midrule
        \textit{Satisfy} [\%]
            & 51.1{\scriptsize$\pm$1.0}
            & 53.6{\scriptsize$\pm$1.0}
            & 56.6{\scriptsize$\pm$1.0}
            & 57.7{\scriptsize$\pm$1.0}
            & \textbf{61.5}{\scriptsize$\pm$0.9} \\
        \bottomrule
    \end{tabular}
    \vspace{\figneg}
    \label{tab:wildbench_pairwise}
\end{table}

\textbf{Main Result.}
\autoref{tab:wildbench_pairwise} reports pairwise win rates and checklist satisfaction rates.
We highlight three patterns.
First, the gains from training with sophisticated user simulators are clear.
The assistant trained with the simplest role-playing LLM (\roleplayone) loses to the assistant trained with \sftmodel 63.9\% of the time,
and it barely beats the initial assistant, winning 55.4\% of the time.
Second, user simulators that incorporate methods to increase realism, from \roleplayone to \roleplaytwo to \roleplaythree,
lead to stronger trained assistants, with higher win rates and checklist satisfaction rates.
However, the assistant trained with \sftmodel still outperforms all assistants trained with role-playing LLMs,
demonstrating that fine-tuning user simulators on human data translates into improved assistants.

\begin{figure}[t]
    \centering
    \includegraphics[width=1.0\linewidth]{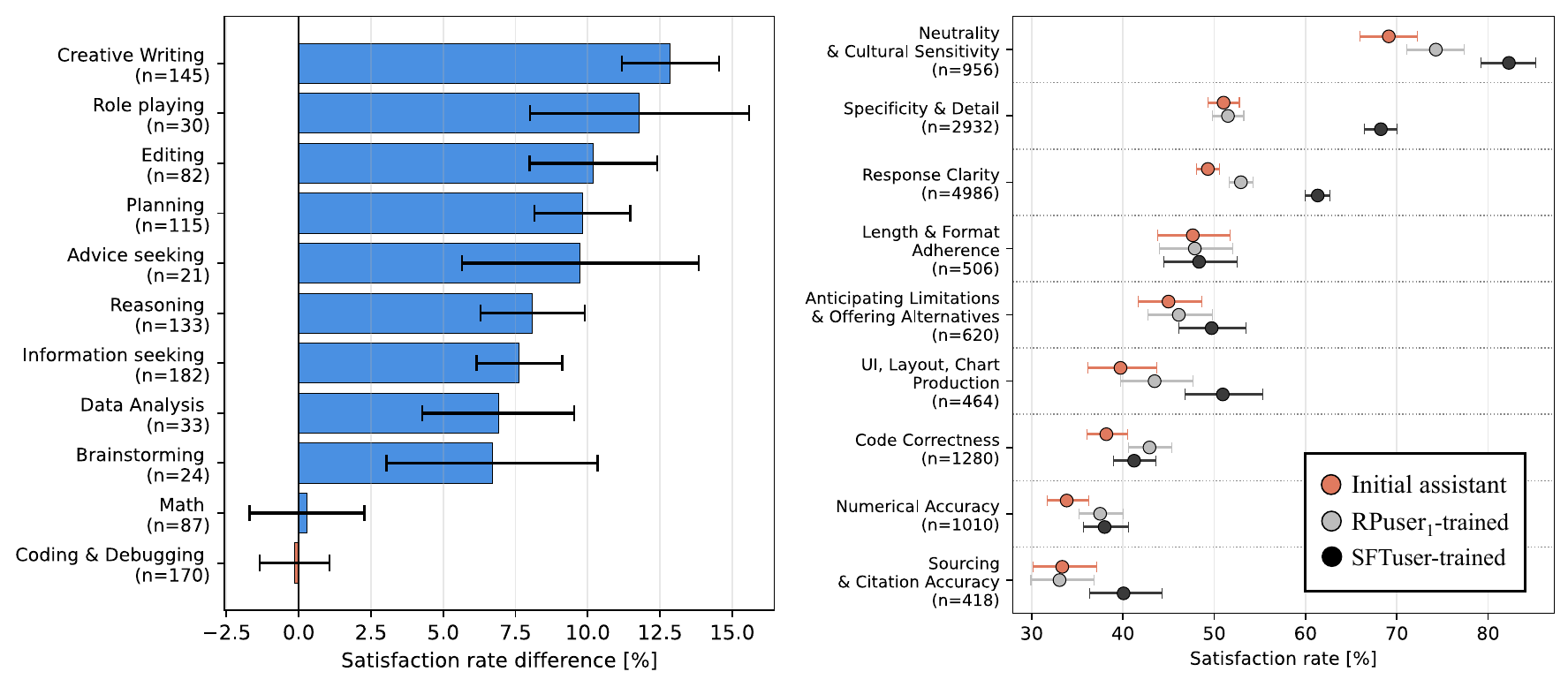}
    \vspace{-20pt}
    \caption{
        \textbf{(Left)}
        Difference in checklist satisfaction rate between the \sftmodel-trained and the \roleplayone-trained assistants, stratified over conversation categories.
        Positive numbers indicate \sftmodel-trained assistant satisfies more items.
        \textbf{(Right)}
        Satisfaction rates among initial, \roleplayone-trained, and \sftmodel-trained assistants,
        stratified over nine representative checklist dimensions.
    }
    \vspace{\figneg}
    \label{fig:intent_categories}
\end{figure}

\textbf{Per-category gap.}
Using WildBench-v2's conversation category taxonomy, we compute the per-category difference in checklist satisfaction rate between the \sftmodel- and \roleplayone-trained assistants (\autoref{fig:intent_categories}, left).
The \sftmodel-trained assistant attains higher rates in most categories,
with the largest gaps in Creative Writing (+12.9\%), Role Playing (+11.8\%), and Editing (+10.2\%).
The exceptions are Math and Coding \& Debugging, where the difference is not statistically significant.
We hypothesize that this pattern reflects the nature of the task:
the two categories with the smallest gaps tend to have a correct answer,
so checklist satisfaction is largely determined by the assistant's problem-solving capability rather than its interaction with the user.
The categories with the largest gaps, by contrast, are open-ended: success depends on accommodating the user's goals over the course of the conversation,
which is precisely what our training signal rewards.

\textbf{Per-dimension gap.}
We also examine the satisfaction rate along finer-grained dimensions to understand what the assistant is learning from simulated conversations.
Starting from all 11,667 checklist items in WildBench,
we apply embedding-based clustering (\autoref{appendix:sec:extended_wildbench}) to extract representative dimensions that items aim to measure (e.g., Numerical Accuracy, Neutrality \& Cultural Sensitivity), then we use gpt-5 to classify whether each item belongs to a given dimension.
We note that this stratification is not exhaustive (e.g., some items are conversation-specific, hence do not belong to any dimensions) but the resulting pattern is informative.
\autoref{fig:intent_categories}, right reports satisfaction rates on nine representative dimensions for the initial, \roleplayone-trained, and \sftmodel-trained assistants.
Compared to the initial assistant, the \sftmodel-trained assistant shows the largest gains on
Neutrality \& Cultural Sensitivity,
Specificity \& Detail,
and Response Clarity, with no significant change on Numerical Accuracy, Code Correctness, and Length \& Format Adherence, 
while the \roleplaytwo-trained assistant only improves significantly on Response Clarity.
Training with \sftmodel thus appears to improve how the assistant engages with the user,
with little effect on its underlying problem-solving capability.
For details about the clustering and representative dimensions, please refer to \autoref{appendix:sec:extended_wildbench}.

\vspace{\myneg}
\subsection{Effects of Scaling User Simulators on Assistant Performance}
\label{sec:experiments:user_sim_metrics}
\vspace{\myneg}

In the previous section, we used Qwen2.5-14B-Instruct as the underlying LLM for all simulators.
Although \roleplayone--\roleplaythree underperformed \sftmodel under this fixed backbone,
it remains unclear whether scaling the underlying LLM would close this gap.
To address this, we conduct an experiment that varies the underlying LLM and measures its effect on the assistant's performance.

\definecolor{heatlo}{HTML}{FADDD3}
\definecolor{heathi}{HTML}{C1E8D5}
\definecolor{bestborder}{HTML}{1D9E75}
\definecolor{cigray}{HTML}{888888}
\definecolor{heatloB}{HTML}{F5C4B3}
\definecolor{heathiB}{HTML}{9FE1CB}
\definecolor{diaggray}{HTML}{EEEEEE}
\definecolor{cigrayB}{HTML}{777777}
\newlength{\tableLmargin}
\newlength{\tableRmargin}
\newlength{\tableGap}
\setlength{\tableLmargin}{-0pt}
\setlength{\tableGap}{0.0em}
\setlength{\tableRmargin}{-0pt}

\vspace{\myneg}
\begin{table}[ht]
\centering
\hspace*{\tableLmargin}
\noindent
\caption{
    WildBench evaluation with scaled user simulators.
    \textbf{(Left)}
    Three assistants trained with \roleplaythree by
    Qwen-2.5-14B-Instruct (\roleplaythree), Qwen3-14B (\roleplayshort-14B), and Qwen3-32B (\roleplayshort-32B).
    \textbf{(Right)}
    Three assistants trained with learned simulators of scales Qwen2.5-\{7,14,32\}B-Instruct.
}
\vspace{-8pt}
\begin{minipage}[t]{0.53\linewidth}
\centering
\small
\renewcommand{\arraystretch}{1.3}
\resizebox{\linewidth}{!}{
\begin{tabular}{r *{5}{w{c}{3.2em}}}
    \toprule
    \textit{Win} [\%]
    & \rot{Initial} & \rot{\roleplaythree} & \rot{\roleplayshort-14B} & \rot{\roleplayshort-32B} & \rot{\sftmodel} \\
    \midrule
    Initial
        & \diagcell
        & \pcell{61.3}{2.7}
        & \pcell{60.0}{2.7}
        & \pcell{63.3}{2.7}
        & \pcellbf{68.4}{2.6} \\
    \roleplaythree
        & \pcell{38.7}{2.7}
        & \diagcell
        & \pcell{48.2}{2.8}
        & \pcell{52.3}{2.8}
        & \pcellbf{58.5}{2.8} \\
    \roleplayshort-14B
        & \pcell{40.0}{2.7}
        & \pcell{51.8}{2.8}
        & \diagcell
        & \pcell{53.3}{2.7}
        & \pcellbf{60.2}{2.7} \\
    \roleplayshort-32B
        & \pcell{36.7}{2.7}
        & \pcell{47.7}{2.8}
        & \pcell{46.7}{2.7}
        & \diagcell
        & \pcellbf{57.4}{2.7} \\
    \midrule
    \textit{Satisfy} [\%]
        & 51.1{\scriptsize$\pm$1.0}
        & 57.7{\scriptsize$\pm$1.0}
        & 57.5{\scriptsize$\pm$1.0}
        & 59.2{\scriptsize$\pm$1.0}
        & \textbf{61.5}{\scriptsize$\pm$0.9} \\
    \bottomrule
\end{tabular}
}
\end{minipage}
\hspace{\tableGap}
\begin{minipage}[t]{0.455\linewidth}
\centering
\small
\renewcommand{\arraystretch}{1.3}
\resizebox{\linewidth}{!}{
\begin{tabular}{r *{4}{w{c}{3.3em}}}
    \toprule
    \textit{Win} [\%]
    & \rot{Initial} & \rot{\sftmodelshort-7B} & \rot{\sftmodel} & \rot{\sftmodelshort-32B} \\
    \midrule
    Initial
        & \diagcell
        & \pcell{67.9}{2.6}
        & \pcell{68.4}{2.6}
        & \pcellbf{75.0}{2.4} \\
    \sftmodelshort-7B
        & \pcell{32.1}{2.6}
        & \diagcell
        & \pcell{51.9}{2.8}
        & \pcellbf{61.5}{2.7} \\
    \sftmodel
        & \pcell{31.6}{2.6}
        & \pcell{48.1}{2.8}
        & \diagcell
        & \pcellbf{56.8}{2.8} \\
    \sftmodelshort-32B
        & \pcell{25.0}{2.4}
        & \pcell{38.5}{2.7}
        & \pcell{43.2}{2.8}
        & \diagcell \\
    \midrule
    \textit{Satisfy} [\%]
        & 51.1{\scriptsize$\pm$1.0}
        & 60.7{\scriptsize$\pm$1.0}
        & 61.5{\scriptsize$\pm$0.9}
        & \textbf{64.8}{\scriptsize$\pm$0.8} \\
    \bottomrule
\end{tabular}
}
\end{minipage}
\hspace*{\tableRmargin}
\label{tab:wildbench_scaling_combined}
\end{table}

\textbf{Scaling Role-Playing LLMs Does Not Yield Stronger Assistants.}
We use Qwen3-\{14,32\}B which substantially outperform Qwen2.5-14B-Instruct on benchmarks measuring LLM capabilities \citep{yang2025qwen3technicalreport}.
We repeat the training with \roleplaythree as a user simulator, which resulted in the best performance among \roleplayone--\roleplaythree (Section~\ref{sec:experiments:wildbench}).
We refer to \roleplaythree with Qwen3-14B as \roleplayshort-14B and \roleplaythree with Qwen3-32B as \roleplayshort-32B.
We train an assistant for each one, all initialized from Qwen2.5-3B-Instruct. 
As shown in \autoref{tab:wildbench_scaling_combined} left, 
scaling the simulator size does not translate into statistically significant gains.
We hypothesize that a stronger LLM produces more fluent utterances, but that increased fluency does not make simulated utterances more representative of actual humans.
This hypothesis aligns with recent findings \citep{zhou2026mind} reporting that stronger instruction-tuned LLMs do not necessarily simulate users more faithfully.

\textbf{Scaling Learned Simulators Yields Stronger Assistants.}
We next ask whether scaling the LLM underlying the learned user simulator translates into stronger assistants.
We prepare learned simulators at three scales by fine-tuning Qwen2.5-\{7,14,32\}B-Instruct on WildChat-1M,
yielding \sftmodelshort-7B, the original \sftmodel (14B), and \sftmodelshort-32B.
We then train three assistants, all initialized from Qwen2.5-3B-Instruct, each paired with one of these simulators.
In contrast to the role-playing case,
scaling the learned simulator yields statistically significant gains in assistant performance (\autoref{tab:wildbench_scaling_combined}, right).
We hypothesize that learned simulators benefit more from scale because
fine-tuning a larger model better captures the distribution of real human utterances and exposes the assistant to that distribution during training.
Consistent with this,
simulator size correlates with lower validation perplexity during \sftmodel training (\autoref{tab:val_perplexity}),
indicating that larger simulators more faithfully imitate real users.

Together, these results show that scaling the simulator's underlying LLM yields statistically significant assistant gains for fine-tuned simulators but not for role-playing ones,
suggesting that the two simulator types differ fundamentally in how they benefit from scale.

\vspace{\myneg}
\subsection{Cross-Simulator Evaluation}
\label{sec:experiments:cross_environment}
\vspace{\myneg}

We now ask a complementary question:
does the assistant remain effective when paired with user simulators it was not trained against?
This matters because an assistant optimized against a single user simulator may learn to exploit that simulator's idiosyncrasies to elicit high reward,
without genuinely improving response quality.
Cross-simulator evaluation lets us distinguish these two cases.
For each pair of simulators $(A,B)$,
we take the assistant trained with $A$ and pair it with $B$ at test time,
conditioning on the fixed intents $z$ from the held-out WildChat-1M test split (Section~\ref{sec:prelim}).

\textbf{Main Result.}
\autoref{tab:figure_cross_environment_eval} reports average test reward.
\emph{Rows are not directly comparable}:
each row corresponds to a different test-time simulator, and simulators differ in behavior---cooperative \roleplayone--\roleplaythree elicit higher judge scores than \sftmodel,
which exhibits behaviors such as information withholding and ambiguity \citep{naous2025flipping}.
Within a row, only the assistant varies, so comparisons are valid.

\definecolor{heatlo}{HTML}{FADDD3}
\definecolor{heathi}{HTML}{C1E8D5}
\definecolor{bestborder}{HTML}{1D9E75}
\definecolor{cigray}{HTML}{888888}

\newcommand{\heatcell}[2]{
    \pgfmathtruncatemacro{\mixpct}{((#1 - 60) / 29) * 100}
    \edef\heatspec{heathi!\mixpct!heatlo}
    \expandafter\cellcolor\expandafter{\heatspec}
    #1\,{\scriptsize\textcolor{cigray}{$\pm$#2}}
}

\newcommand{\heatcellbest}[2]{
    \pgfmathtruncatemacro{\mixpct}{((#1 - 60) / 29) * 100}
    \edef\heatspec{heathi!\mixpct!heatlo}
    \expandafter\cellcolor\expandafter{\heatspec}
    \textbf{#1}\,{\scriptsize\textcolor{cigray}{$\pm$#2}}
}

\newcommand{\heatcellrunner}[2]{
    \pgfmathtruncatemacro{\mixpct}{((#1 - 60) / 29) * 100}
    \edef\heatspec{heathi!\mixpct!heatlo}
    \expandafter\cellcolor\expandafter{\heatspec}
    \underline{#1}\,{\scriptsize\textcolor{cigray}{$\pm$#2}}
}

\begin{table}[t]
    \centering
    \small
    \setlength{\tabcolsep}{4pt}
    \renewcommand{\arraystretch}{1.3}
    \caption{
        Cross-simulator evaluation of trained assistants.
        Rows correspond to user simulators used at test time;
        columns are assistants trained with different user simulators.
        We normalize judge scores (Section~\ref{sec:experiments:setup}) from 0--10 to 0--100 for convenience.
        In each row, the best and the runner-up are bold-faced and underlined, respectively; {\scriptsize$\pm$} indicates 95\% CI from bootstrapping scores.
    }
    \begin{tabular}{r ccccc}
        \toprule
        & & \multicolumn{4}{c}{\textbf{Assistant trained with}} \\
        \cmidrule(lr){3-6}
        \textbf{Eval with}
            & Initial
            & \roleplayone
            & \roleplaytwo
            & \roleplaythree
            & \sftmodel \\
        \midrule
        \roleplayone
            & \heatcell{68.5}{0.4}
            & \heatcellrunner{88.5}{0.2}
            & \heatcell{87.6}{0.2}
            & \heatcell{87.6}{0.2}
            & \heatcellbest{88.7}{0.2} \\
        \roleplaytwo
            & \heatcell{63.3}{0.4}
            & \heatcell{82.9}{0.2}
            & \heatcellrunner{86.8}{0.2}
            & \heatcell{86.3}{0.2}
            & \heatcellbest{87.5}{0.2} \\
        \roleplaythree
            & \heatcell{63.1}{0.4}
            & \heatcell{79.8}{0.3}
            & \heatcell{82.0}{0.3}
            & \heatcellrunner{83.8}{0.2}
            & \heatcellbest{84.8}{0.2} \\
        \sftmodel
            & \heatcell{60.9}{0.4}
            & \heatcell{72.9}{0.4}
            & \heatcell{78.2}{0.3}
            & \heatcellrunner{80.3}{0.3}
            & \heatcellbest{82.5}{0.3} \\
        \midrule
        \sftmodeltwo
            & \heatcell{61.1}{0.4}
            & \heatcell{73.0}{0.3}
            & \heatcell{74.7}{0.3}
            & \heatcellrunner{76.8}{0.3}
            & \heatcellbest{82.8}{0.2} \\
        \bottomrule
    \vspace{\figneg}
    \vspace{\figneg}    
    \end{tabular}
    \label{tab:figure_cross_environment_eval}
\end{table}

\begin{figure}[t]
    \centering
    \includegraphics[width=1.00\linewidth]{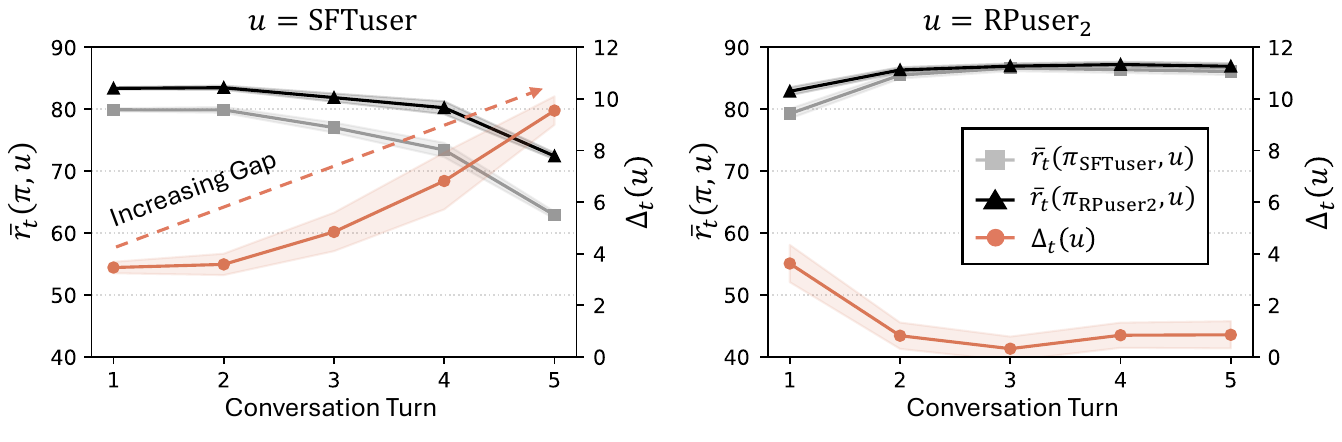}
    \vspace{-20pt}
    \caption{
    Per-turn mean reward $\bar{r}_t(\pi,u)$ and per-turn reward difference $\Delta_t(u)$ between the \sftmodel- and \roleplaytwo-trained assistants
    \textbf{(Left)} evaluated with \sftmodel and \textbf{(Right)} \roleplaytwo.
    Shaded regions denote $\pm 1$ standard error.
    With \sftmodel, the gap widens with increasing turn depth.
    }
    \vspace{\figneg}
    \label{fig:figure_diff}
\end{figure}

The \sftmodel-trained assistant achieves the highest reward in every row, including rows for simulators it did not see during training. Assistants trained with role-playing LLMs,
by contrast, are competitive only with their train-time partner.
Their performance gap to the \sftmodel-trained assistant is small ($\leq 1.0$) when tested against role-playing LLMs but widens to $2.2$ against \sftmodel.
Pairing each assistant with \sftmodeltwo, a held-out simulator fine-tuned on data distinct from \sftmodel (Section~\ref{sec:user_simulators:sft}),
yields the same conclusion:
the \sftmodel-trained assistant remains the strongest (82.8 vs.\ 76.8 for the runner-up).

These results support two related hypotheses.
First, the learned user simulator exposes the assistant to a broader distribution of user behaviors,
producing an assistant that generalizes across simulators \citep{peng2018sim}.
Second, the strong train-time performance of role-play-trained assistants reflects
optimization against a narrow band of simulator-specific behaviors rather than improved interaction/response quality \citep{carroll2019utility}.

\textbf{Per-turn performance gap.}
Now, we ask a finer-grained question: does the gap between assistants grow with the conversation length?
If so, the asymmetry observed in \autoref{tab:figure_cross_environment_eval} would compound rather than remain constant.
We compare two assistants, the one trained with \sftmodel and the one trained with \roleplaytwo.
We define the per-turn performance gap when both are paired with a user simulator $u$ at test time as $\Delta_t(u) = \bar{r}_t(\pi_{\textsc{SFTuser}},\, u) - \bar{r}_t(\pi_{\textsc{RPuser2}},\, u)$,
where $\bar{r}_t(\pi_{u'},u)$ denotes the mean reward over conversations of length $t$
when the assistant trained with $u'$ is paired with $u$.

\autoref{fig:figure_diff} plots $\Delta_t(u)$ for $u{=}$\sftmodel (left) and $u{=}$\roleplaytwo (right).
When tested against \roleplaytwo, both $\bar{r}_t$ and $\Delta_t$ remain approximately flat,
as \roleplaytwo's structured and cooperative follow-ups (Section~\ref{sec:prelim}) pose little additional challenge as the conversation extends.
When tested against \sftmodel, however, $\Delta_t$ grows with turn depth:
rewards of both assistants decrease at later turns, reflecting the increasing difficulty of \sftmodel's follow-ups,
but the decline is markedly steeper for the \roleplaytwo-trained assistant.
The \roleplaytwo-trained assistant has not encountered such follow-ups from \sftmodel during training,
so its performance degrades as conversations with \sftmodel extend — analogous to compounding policy error under distributional shift \citep{janner2019trust, feinberg2018model}.

\textbf{Relation to prior work.}
These results are compatible with prior simulation-based assistant training works yet reveal their unseen limitation \citep{kong2024platolm, wu2025collabllm, dinucu2025problem, qian2025userrl, li2024iqa, luo2024duetsim, gao2024regressing, shani2024multi}.
Prior works evaluate trained assistants in the matched setting,
pairing an assistant trained with a role-playing LLM against the same role-playing LLM at test time.
Our cross-simulator results are consistent with theirs in that matched regime,
but reveal a limitation invisible to it:
high matched-setting performance can coexist with brittle generalization to different types of user simulators.
\section{Real-World User Study}
\label{sec:user_study}
\vspace{\myneg}

\begin{figure}[t]
    \centering
    \includegraphics[width=1.0\linewidth]{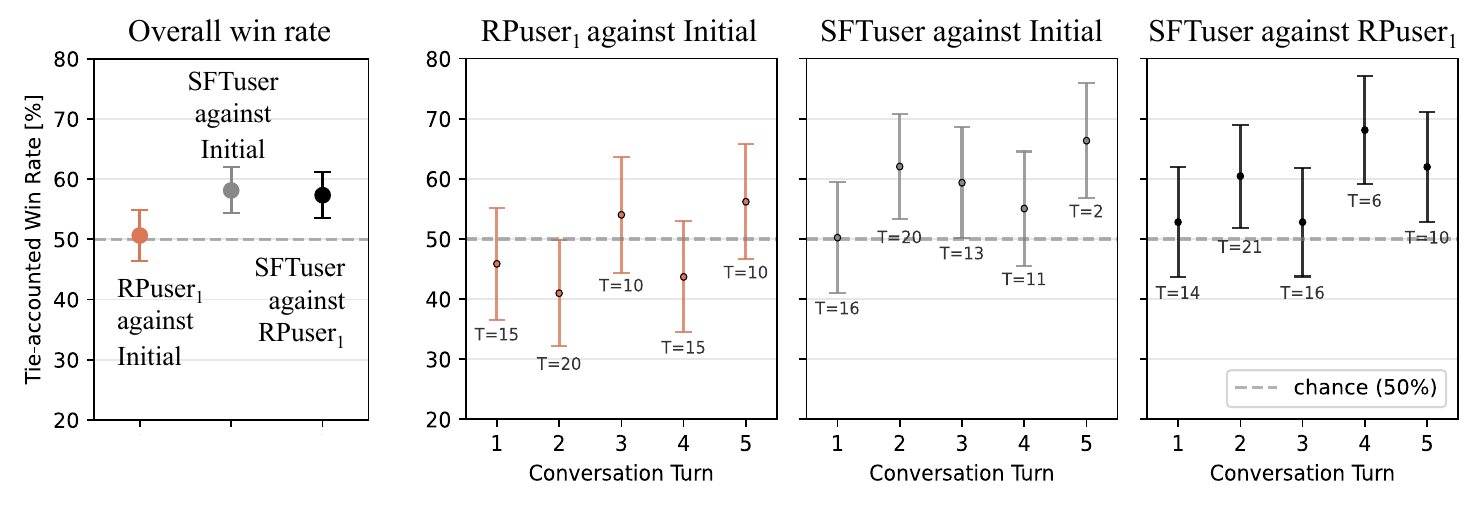}
    \vspace{-20pt}
    \caption{
    Tie-accounted pairwise win rates from the real-world user study ($n=283$, $\geq 5$ turns each).
    \textbf{(Left)}
    Win rates for the three assistant pairs, aggregated across turns. 
    \textbf{(Right)}
    Win rates stratified by turn.
    Error bars denote 95\% conversation-level bootstrapped confidence intervals \citep{davison1997bootstrap};
    `$T{=}\cdot$' below each error bar indicates the number of ties
    (out of $\geq 90$ participants per assistant pair). 
    }
    \vspace{\figneg}
    \label{fig:figure_human_study}
\end{figure}

We conduct a user study on Prolific with 283 participants, following the protocol of \citet{wu2025collabllm}.
Each participant is randomly assigned one of three writing tasks---blog post, creative writing, or personal statement---and selects an intent from a curated list
(e.g., ``Write a blog post about taking up a new language as an adult.'').
We focus on writing because it avoids the personally identifiable information that arises in fully open-domain settings (e.g., rewriting personal emails) while remaining accessible to a general population, unlike specialized tasks such as Python coding.
We allow participants to select their intent so that they can write about something they are familiar with, as a better proxy for a real-world user who genuinely has this intent.
Towards this end, we also have participants answer a short set of pre-writing questions before starting (e.g., ``Who is the audience for this blog post?'').

After a single-turn practice session to familiarize themselves with the study interface,
participants engage in at least five turns of conversation with two anonymized assistants,
sampled from \{Initial, \sftmodel-trained, \roleplayone-trained\}, whose responses are displayed side by side in randomized positions.
At each turn, the participant selects the preferred response and provides a brief explanation guided by the rubric; the conversation continues from the chosen response.
Full details of the study design, including recruitment, a set of writing topics, and quality-control measures, are in \autoref{appendix:sec:human_study_detail}.

\textbf{Quantitative results.}
\autoref{fig:figure_human_study} reports pairwise win rates for the three assistant pairs, aggregated across turns (left) and stratified by turn (right).
Each pair is evaluated by at least 90 participants over at least 5 turns, yielding at least 450 pairwise decisions per pair and at least 90 per turn.

In aggregate, the \sftmodel-trained assistant outperforms both the initial assistant (58.1\%, CI [54.3, 61.8]) and the \roleplayone-trained assistant (57.3\%, CI [53.4, 61.1]).
In contrast, the \roleplayone-trained assistant is statistically indistinguishable from the initial assistant (50.6\%, CI [46.5, 54.9]).
While the exact numbers differ from WildBench, which tested a different set of tasks, overall conclusions remain similar: on WildBench, training with \roleplayone also only yielded marginal gains over the initial assistant (55.4\%, CI [52.7, 58.1]; Section~\ref{sec:experiments:wildbench}), while the \sftmodel-trained assistant significantly outperformed both the initial and \roleplayone-trained assistants.
When stratified by turn, confidence intervals widen due to smaller samples, but the aggregate pattern persists.
The \sftmodel-trained assistant's estimated win rate stays above 50\% at every turn against both comparators,
while the \roleplayone-trained assistant oscillates around 50\% against the initial assistant.
We also observe that the \sftmodel-trained assistant's win rate tends to increase with turn depth,
consistent with the per-turn trend observed in cross-simulator multi-turn evaluation (Section~\ref{sec:experiments:cross_environment}).

\textbf{Qualitative results.}
We use these participants' choice explanations to characterize each assistant's perceived strengths and weaknesses;
representative examples appear in \autoref{tab:human_study_rationales}.
The \sftmodel-trained assistant is frequently described as personal, grounded, and context-aware (``comprehensive and organized,'' ``feels natural,'' ``more engaged in the conversation''), but loses on prompts demanding brevity or strict format compliance, where some raters find it too long or too literary.
The \roleplayone-trained assistant, conversely, is preferred for thoroughness and surface polish---complete drafts and clean structure---yet is often flagged as stiff and bullet-heavy.
The initial assistant is concise and direct, which raters value on short-form prompts, though some note that it can lose track of earlier-turn details as conversations deepen.
Taken together, these explanations complement the quantitative findings:
training with a learned user simulator yields an assistant whose outputs are perceived as more engaging and context-aware in multi-turn conversations.

\begin{table}[t]
\centering
\small
\caption{
    Example explanations from human participants for their pairwise decisions from the user study,
    reproduced verbatim (grammatical and spelling errors are preserved).
}
\label{tab:human_study_rationales}
\begin{tabular}{@{}p{0.155\linewidth}p{0.085\linewidth}p{0.70\linewidth}@{}}
\toprule
\textbf{Assistant Pair} & \textbf{Choice} & \textbf{Explanation of Choice} \\
\midrule
\multirow{2}{=}{\roleplayone-trained \\ vs.\ Initial}
  & \roleplayone-trained 
    & ``though both were able to complete the prompt , it seemed like a was able to create a slightly longer letter which is what i was aiming for.'' \\
\cmidrule(l){2-3}
  & Initial Assistant
    & ``di better this time because the responses are straight to the 
       point, which is better for a short blog post'' \\
\midrule
\multirow{2}{=}{\sftmodel-trained \\ vs.\ Initial}
  & \sftmodel-trained 
    & ``It was more comprehensiveness and organized, it divided the statement 
       into obvious sections with bolded headlines which help the reader to follow up \ldots'' \\
\cmidrule(l){2-3}
  & Initial Assistant
    & ``I think it better due to the simplicity and relatable nature of 
       many of the steps, with the addition of providing many out of the 
       box options \ldots'' \\
\midrule
\multirow{2}{=}{\sftmodel-trained \\ vs.\ \roleplayone-trained}
  & \sftmodel-trained 
    & ``It feels more personal and natural, with better reflection on 
       challenges, growth, and motivation. Other sounds more generic and 
       repetitive.'' \\
\cmidrule(l){2-3}
  & \roleplayone-trained 
    & ``Other includes unnecessary titles `First Paragraph, Second Paragraph'.
        It is clear and simple although it will need one more round, titles could be better.'' \\
\bottomrule
\end{tabular}
\end{table}
\section{Related Work}
\label{sec:related_work}
\vspace{\myneg}

In the Extended Related Work (\autoref{appendix:sec:extended_related_work}),
we also discuss the sim-to-real gap in RL, multi-turn preference optimization, rubric-as-a-reward, and the reward overoptimization phenomenon.

\textbf{User Simulation with LLMs.}
A body of work investigates using language models to simulate human users.
One approach prompts an instruction-tuned LLM to role-play as a user in a zero-shot or few-shot fashion.
In task-oriented dialogue, \citet{terragni2023context} and \citet{davidson2023user} build user simulators via in-context learning,
while \citet{luo2024duetsim} employs two LLMs (generator and verifier) to improve goal adherence.
Beyond dialogue systems,
\citet{zhang2025llm} uses prompted LLMs to simulate user engagement for recommender systems,
and \citet{tamoyan2025llm} constructs persona-based prompted simulators for demographically diverse chatbot interactions.
Recently, \citet{shim2025non} develops user simulators with a scaffold to test agents under adversarial user behaviors.

However, a growing line of work reveals that instruction-tuned LLMs are poor proxies for human users:
they produce predictable, cooperative, and well-structured utterances not stylistically aligned with humans \citep{ivey2024robotic, yoon2024evaluating, zhou2026mind}.
These findings motivate approaches that instead train language models on human data.
In a conversational setting, \citet{naous2025flipping} introduces UserLM,
a model fine-tuned on the user side of WildChat conversations \citep{zhao2024wildchat}, producing more realistic user utterances than prompted alternatives.
\citet{wang2025know} and \citet{wu2026humanlm} extend this by training LLMs to generate internal chain-of-thought reasoning;
\citet{gandhi2026learning} pursues a related objective of simulating human-human dialogues;
\citet{ccalik2025enhancing} applies DPO \citep{rafailov2023direct} to synthetic pairwise data that up-votes human-like generations.
While prior works focus on building and evaluating user simulators or evaluating LLM assistants against them,
ours is the first to systematically study how the choice of LLM-based user simulator affects the trained assistant's capability to collaborate with humans.

\textbf{Training LLM Assistants with User Simulators.}
Recent works train LLM assistants via interaction with simulated users.
On the algorithmic side, several works pursue multi-turn preference optimization \citep{shani2024multi, shi2024direct};
for instance, REFUEL \citep{gao2024regressing} is an on-policy method that regresses the log-policy-ratio towards the relative Q-value obtained via MC estimation.
In the human--LLM interaction domain, prior works target the collaboration capabilities of LLMs.
This includes clarifying-question generation by simulating user interactions \citep{andukuri2024star, zhang2024modeling},
multi-turn collaboration \citep{wu2025collabllm},
proactive information elicitation \citep{kong2024platolm},
and pedagogical capability via interaction with an LLM role-playing student \citep{dinucu2025problem}.
In spite of the growing body of work on building more realistic user simulators,
a consistent thread in training LLM assistants via simulation is the use of role-playing LLMs.
\section{Conclusion, Limitations, And Future Work}
\label{sec:limitations}
\vspace{\myneg}

In this work, we reframed simulator quality in terms of its downstream consequences: how an LLM assistant trained with the user simulator performs with human users.
We showed that training assistants with role-playing LLMs yields only negligible downstream improvements, while substantive gains come from fine-tuning simulators on human data.
Furthermore, improvements to a simulator do not always transfer to its assistant:
scaling the model underlying a role-playing simulator does not translate into improved assistant performance,
and techniques such as persona conditioning yield incremental gains but do not close the gap to the assistant trained with \sftmodel.
These results are consistent across WildBench, cross-user evaluation, and a real-world user study.

\textbf{Limitations and future work.}
Our framework also surfaces limitations that could be explored in future work.
First, while training against user simulators can be effective, the simulator must remain realistic as the assistant's policy shifts during training.
Second, we focus on open-domain conversation, given its generality and the availability of data,
but future work could explore specialized environments such as $\tau$-bench~\citep{yao2024tau}, where 
dialogues are goal-oriented and conditioned on long tool-augmented contexts.
We discuss concrete mitigations for these issues in Appendix~\ref{sec:appendix:extended_limitations},
including bounded-horizon rollouts branched from real conversation prefixes \citep{feinberg2018model, sutton1991dyna},
uncertainty-aware or iteratively retrained user simulators \citep{ross2011reduction, chua2018deep},
and collecting human interaction data across a broader range of domains \citep{shim2025non, zhou2026mind}.
Our work also establishes the importance of evaluating user simulators downstream---based on the performance of their trained assistants with real humans---but this metric is expensive to measure, requiring a user study in the best case or at least a real-world task benchmark like WildBench.
Future work could explore accurate upstream proxies of this metric, enabling more efficient iteration on simulator design.

\begin{ack}
We are grateful to Prof. John Canny, Prof. Emma Pierson, David Chan, Suhong Moon, Philippe Laban, and Sehoon Kim for their thoughtful comments and valuable feedback.
We thank Jonathan Ngai, Murari Ganesan, Steven Luo, Erfan Jahanparast, Carlos Guirado, and Jessie Li for their assistance with the pilot user study.
Joseph Suh acknowledges support from the Korea Foundation for Advanced Studies through the Doctoral Study Abroad Program. This work was supported in part by the Google Research Scholar Program and by compute resources from the Center for Human-Compatible AI (CHAI) and VESSL AI.
The views and findings expressed herein are those of the authors and do not reflect the official positions or policies of any sponsoring organization.
\end{ack}

\clearpage
\bibliographystyle{unsrtnat}
\bibliography{reference}

\appendix
\section{Extended Related Work}
\label{appendix:sec:extended_related_work}
\vspace{\myneg}

\textbf{Sim-to-Real Transfer in RL.}
A challenge in RL is that an agent trained in a simulated environment overfits to its specific dynamics, developing a brittle policy that fails under distributional shift \citep{cobbe2019quantifying}.
Domain randomization \citep{tobin2017domain, mehta2020active} addresses this by training an agent across simulation environment parameters (e.g., friction coefficient and visual appearance) so that the policy learns to be robust to variation \citep{peng2018sim, andrychowicz2020learning}.
In a similar spirit, a related strategy in multi-agent settings is to diversify partners an agent trains with.
Population-based training (PBT) \citep{jaderberg2019human, zhao2023maximum} maintains a population of co-evolving agents, exposing the agent to diverse partner behaviors.
However, PBT agents may co-adapt and converge on opaque coordination conventions that fail when paired with humans \citep{carroll2019utility}.

\citet{carroll2019utility} demonstrates that training an agent against a behavior-cloned human model in a cooperative game (Overcooked) produces a policy that coordinates with actual humans.
Our work builds on this insight.
By preparing user models via supervised fine-tuning on human user utterances \citep{naous2025flipping, meshi2026convapparel},
we construct a multi-turn conversation environment with a learned user simulator.
Each rollout against this simulator effectively samples from a space of human utterances,
varying in belief, desire, intent, and linguistic patterns \citep{kirk2024prism, herlihy2024overcoming, yang2025prompts}.
This training environment produces agents that are robust to the wide range of actual users,
in line with the broader effort to train under greater environmental variability for transferrable policies \citep{pinto2017robust, ghosh2021why, chen2021understanding}.

\textbf{Rubrics as Rewards and Reward Overoptimization.}
Many real-world tasks, including successful open-ended conversation,
require nuanced and multi-criteria evaluation beyond a verifiable reward.
Recent works tackle this problem by decomposing desired response qualities into rubrics
and using LLM judges to score each criterion \citep{gunjal2025rubrics, huang2025reinforcement}.
For example, \citet{viswanathan2025checklists} proposes instruction-specific checklists verified by both AI judges and programmatic verifiers;
Kimi K2 \citep{team2025kimi} combines RLVR with a self-critique rubric reward.
We adopt a rubric-based reward in a similar way:
we define an evaluation criteria and score assistant responses in simulated conversations via LLM.

However, optimizing against any proxy reward carries the risk of reward overoptimization
\citep{gao2023scaling, rafailov2024scaling}.
When the reward comes from an LLM judge,
the risk is compounded by judge biases such as verbosity preference and self-preference \citep{zheng2023judging}.
For example, LLMs optimized for simulated user feedback can learn targeted manipulation and deception \citep{williams2024targeted}.
In our setting,
overoptimization could manifest as the assistant learning to produce responses that game the judge to get higher scores without genuine quality improvement.
We monitor by evaluating against multiple LLM judges \citep{coste2023reward, verga2024replacing, gu2024survey}
and by using a disjoint set of LLM judges between train and test time.

\textbf{Single-Turn Limitations of Standard RLHF.}
Several recent works address the single-turn limitations of RLHF by extending to multi-turn settings.
\citet{shani2024multi} introduces Multi-Turn Preference Optimization, which solicits preference feedback over entire conversations rather than individual turns, proving convergence to Nash equilibrium in the tabular setting.
\citet{gao2024regressing} proposes REFUEL, which casts multi-turn RLHF as a Markov decision process and performs iterative on-policy data collection.
Our work leverages this insight: by training agents via on-policy multi-turn rollouts against a user model,
the agent learns from conversation trajectories it generates itself and optimizes for long-horizon outcomes.
However, we focus on investigating the utility of learned user model rather than relying on the completely synthetic human--LLM conversation data with role-playing assistant LMs as previous multi-turn preference optimization works have done.
\vspace{\myneg}
\section{Extended Discussion of Limitations and Future Work}
\label{sec:appendix:extended_limitations}
\vspace{\myneg}

\textbf{Learned User Simulator Faces Distributional Shift.}
\sftmodel learns to imitate human utterances conditioned on conversation histories present in the training data,
but faces distributional shift when conditioned on out-of-distribution histories --- the histories that a changing assistant policy will produce.
As the policy changes during training, it may steer conversation trajectories into regions of the state space that \sftmodel was not trained on,
causing \sftmodel to produce unrealistic utterances and allowing the policy to exploit inaccurate transition dynamics \citep{deisenroth2011pilco}.

One direction to mitigate this out-of-distribution model behavior is to limit the horizon of simulated trajectories.
In a model-based RL,
\citet{feinberg2018model} demonstrates that branching short model-based rollouts from states visited in real world controls the compounding model error.
This approach uses the model's local accuracy while staying grounded in the real state distribution \citep{sutton1991dyna}.
An analogous strategy in our setting is to start each simulated conversation from a real conversation prefix, and to rollout only a small number of turns under the user simulator.
This limits the depth over which user simulator errors can compound, while still allowing the assistant to learn from multi-turn consequences of its actions.
We already implement a limited form of this idea in Section~\ref{sec:user_simulators:sft}:
we ground the initial utterance $o_1$ in real queries and uses \sftmodel for subsequent transitions.
The natural extension is to branch from real conversation prefixes at every turn, not only the first.

This direction introduces a tradeoff.
Shorter rollout windows reduce user simulator error accumulation but also limit the temporal horizon over which the assistant can learn to plan.
If the assistant only observes one or two simulated future turns,
it may fail to develop strategies (e.g., asking clarifying questions) that pay off over longer horizons.
Characterizing this tradeoff, and developing adaptive rollout length schemes (e.g., \citet{janner2019trust}), is an important direction for future work.

A complementary approach is an uncertainty-aware user simulator \citep{chua2018deep, kurutach2018model} or iteratively improve the user simulator.
If the trained assistant is iteratively deployed to collect fresh human interaction data,
the user simulator can be retrained to cover the new regions of state space that the updated assistant policy induces.
This data aggregation strategy \citep{ross2011reduction} directly addresses distributional shift, though it reintroduces the cost of human data collection that simulation was designed to avoid.
Hybrid schemes of occasional human data collection interleaved with simulation may offer a practical middle ground.

\textbf{Generalization to Other Simulation Domains.}
Many LLM agent training environments rely on domain-specific user simulators.
A prominent example is $\tau$-bench~\citep{yao2024tau},
whose user simulator is a role-playing instruction-tuned LLM---the type of simulator that recent work has shown to produce unrealistic transition dynamics~\citep{shim2025non}.

Applying a user simulator fine-tuned on WildChat to $\tau$-bench is appealing but faces obstacles.
First, there is a domain gap:
$\tau$-bench conversations are goal-oriented dialogues with structured task constraints,
whereas \sftmodel (Section~\ref{sec:user_simulators:sft}) is trained on open-domain human--LLM conversations.
The pragmatics of a human seeking agent support in $\tau$-bench
(e.g., concisely describing a problem, responding to diagnostic questions)
may differ substantially from those of open-ended conversation.
Second, the $\tau$-bench user simulator operates in a long, tool-augmented context.
It is conditioned not only on the conversation history but also on a specification of available actions and tool signatures spanning thousands of tokens.
\sftmodel is trained to generate utterances conditioned only on a high-level intent $z$ and a conversation history, hence struggles to follow such long context instructions.

Despite these challenges, the motivation of our work---user simulators grounded in human data produce more realistic transition dynamics than role-playing LLMs---is
not specific to open-domain conversation.
\citet{zhou2026mind} collected a corpus of human--agent interactions in $\tau$-bench setting,
providing an initial step toward data collection in specialized domains.
We hope that the growing recognition of the inadequacy of role-playing user simulators~\citep{shim2025non, naous2025flipping, zhou2026mind}
motivates the community to collect human interaction data across a broader range of domains,
enabling domain-specific user simulators that provide faithful simulation environments.

\textbf{Language coverage.}
All trained and evaluated assistants and corpora are English-language.
Whether the findings of the work hold in other languages is an open empirical question.
\vspace{\myneg}
\section{Supervised Fine-Tuning On User-side Utterances}
\label{appendix:sec:behavioral_cloning}
\vspace{\myneg}

We outline the data preprocessing steps for WildChat-1M and WildChat-4.8M,
and describe the supervised fine-tuning hyperparameters and checkpoint selection to prepare \sftmodel and \sftmodeltwo (Section~\ref{sec:user_simulators:sft}).
We end this section with input--output specifications of the learned user simulator.

\vspace{\myneg}
\subsection{WildChat-1M}
\label{sec:appendix:wildchat}
\vspace{\myneg}

We begin with allenai/WildChat-1M, a collection of 838K conversations between human users and GPT-family models (\texttt{gpt-3.5-turbo} or \texttt{gpt-4}), derived from allenai/WildChat-1M-Full by removing toxic conversations flagged by OpenAI Moderations API or Detoxify \citep{zhao2024wildchat}.
We apply the following exclusion criteria sequentially:
\begin{enumerate}
    \item \textbf{Deduplication.} We remove conversations with duplicate \texttt{conversation\_hash}, as these contain identical content. This reduces the corpus from 837,989 to 826,319 conversations.
    \item \textbf{Language filtering.} We retain only English conversations, yielding 474,405 conversations.
    \item \textbf{$n$-gram filtering.} We remove any conversation in which the first user turn contains at least one 7-gram that appears more than 100 times across the entire corpus. This step filters out formulaic queries that users frequently copy and paste into the data collection platform, reducing the corpus to 301,396 conversations.
\end{enumerate}

For each remaining conversation, we generate a high-level user intent summary, following \citet{naous2025flipping}.
Specifically, we provide the full human--LLM conversation trajectory to Qwen/Qwen3-32B (thinking enabled) and instruct it to produce a short summary of the user's high-level intent.
We refer the reader to the ``Prompt Template for Intent Generation'' in \citet{naous2025flipping} for the exact prompt.

Followed by intent generation, we partition the dataset into train, validation, and test splits with a ratio of 89\%\,/\,5\%\,/\,6\% at a \emph{user} level.
WildChat conversations contain \texttt{hashed\_ip} field that encrypts the IP address of user; we use this field to ensure that all conversations from a single human user belong to one split.
This results in 268,636 train, 14,142 validation, and 18,618 test conversations.

\vspace{\myneg}
\subsection{WildChat-4.8M}
\vspace{\myneg}

We additionally use allenai/WildChat-4.8M, a collection of 3.20M conversations between human users and GPT-family models, derived from allenai/WildChat-4.8M-Full by removing 1.54M toxic conversations~\citep{deng2024wildvis}.
While multiple GPT models are used for data collection,
\texttt{gpt-4o} and \texttt{gpt-4.1-mini} constitute 90\% of conversations.
We apply all exclusion criteria for WildChat-1M, with an additional step to remove an overlap between the two datasets:
\begin{enumerate}
    \item \textbf{Deduplication.} We remove conversations with duplicate \texttt{conversation\_hash}, reducing the corpus from 3,199,860 to 2,869,082 conversations.
    \item \textbf{Overlap removal.} Of the remaining conversations, 758,979 share \texttt{conversation\_hash} with an entry in WildChat-1M. We verified that conversations sharing a hash across the two datasets are identical. Removing these yields 2,110,103 conversations.
    \item \textbf{Language filtering.} We retain only English conversations, yielding 1,016,665 conversations.
    \item \textbf{$n$-gram filtering.} Applying the same 7-gram filtering criterion as above reduces the corpus from 1,016,665 to 247,830 conversations. The substantial drop of 768,835 conversations indicates a high prevalence of frequently repeated queries in this larger collection.
\end{enumerate}

For each remaining conversation, we perform the identical intent generation process followed by the user-level split, yielding 224,501 train, 10,412 validation, and 12,917 test conversations.

\vspace{\myneg}
\subsection{Supervised Fine-Tuning}
\vspace{\myneg}

We perform supervised fine-tuning (SFT) on the user-side utterances of the training split, using a custom FSDP2 + torch.compile implementation built on top of the llama-cookbook repository \citep{llamacookbook2025}.
Below, we present a training sample from WildChat-1M formatted with the Qwen2.5 chat template, where \hl{highlighted spans} indicate the tokens over which the cross-entropy loss is computed:

{
\small
\noindent
\begin{tabular}{@{}p{\textwidth}@{}}
\texttt{<|im\_start|>system} \\
\texttt{You are a user chatting with an assistant language model to convert a website into a PDF format.<|im\_end|>} \\
\texttt{<|im\_start|>user} \\
\hl{\texttt{To make the following website as an PDF<|im\_end|>}} \\
\texttt{<|im\_start|>assistant} \\
\texttt{To convert a website to a PDF, you can follow these steps: \ldots (truncated) \ldots but not its dynamic or interactive elements.<|im\_end|>} \\
\texttt{<|im\_start|>user} \\
\hl{\texttt{Can you do that for me?<|im\_end|>}} \\
\texttt{<|im\_start|>assistant} \\
\texttt{Apologies for the confusion, but as a text-based AI, \ldots (truncated) \ldots to assist you.<|im\_end|>} \\
\texttt{<|im\_start|>user} \\
\hl{\texttt{<|endconversation|><|im\_end|>}} \\
\texttt{<|endoftext|>}
\end{tabular}
}

We explain some details.
First, the prepared intent (``You are a user chatting\ldots'') is placed in the system prompt to condition generation.
Second, the loss is applied not only to user-turn content but also to the turn-ending special token (\texttt{<|im\_end|>} for Qwen2.5-family models), so that the model learns to produce delimited turns.
Third, following \citet{naous2025flipping}, we introduce a special token \texttt{<|endconversation|>} to signal that the user has terminated the conversation:
a real human--LLM conversation with $2n$ turns is structured into $2n+1$ turns with $n+1$ user-side utterances (the last one always being \texttt{<|endconversation|>}) and $n$ LLM responses.
We initialize the embedding of the special token with the pretrained EOS token embedding.

\textbf{Sequence packing.}
Conversation samples exhibit high length variance, so we employ sequence packing with a maximum sequence length of 16,384 tokens to improve training throughput.
Within a packed sequence, individual samples are separated by the padding token (\texttt{<|endoftext|>} for Qwen2.5-family models), and we ensure that no single sample is split across packed sequences.
Explicit masking of cross-sample attention via block-diagonal masks is a viable strategy not taken here.
This packing scheme has two consequences:
(1)~conversations exceeding 16,384 tokens are dropped, which affects fewer than 1\% of the training samples; and
(2)~packed sequences vary in length, so we apply left-padding to produce uniform-length sequences compatible with the static memory layout required by torch.compile.

\textbf{Hyperparameters and compute hours.}
We perform full fine-tuning, updating all parameters including the embedding and the language modeling head.
We use the Adam optimizer~\citep{kingma2014adam} with a linear-warmup-cosine-decay learning rate schedule:
the learning rate warms up over 10\% of total steps from 0 to a peak of $2 \times 10^{-5}$, then decays to $2 \times 10^{-6}$.
Training runs for 2 epochs with a batch size of 64 packed sequences, on 8 A100-SXM4-80GB GPUs with per-GPU micro batch size of 1 and gradient accumulation.
It takes 120 (260) GPU-hours for 14B (32B) model fine-tuning.

\begin{figure}[t]
    \centering
    \includegraphics[width=0.8\linewidth]{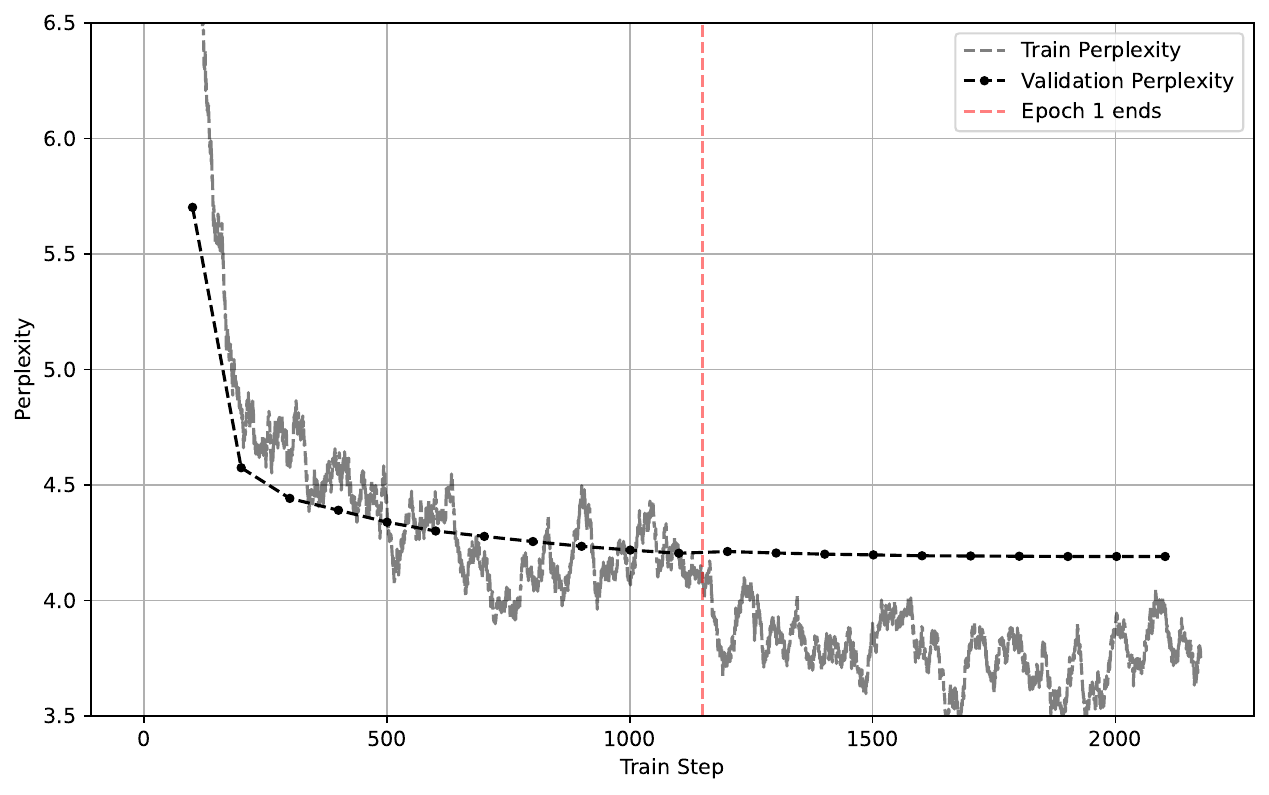}
    \caption{
    Training curve of Qwen/Qwen2.5-14B-Instruct on WildChat-1M (\sftmodel).
    \vspace{-10pt}
    }
    \label{fig:figure_sft_loss_curve}
\end{figure}

\textbf{Checkpoint selection.}
We evaluate on the validation set every 100 training steps (i.e., consuming 6,400 packed sequences) and select the checkpoint with the lowest corpus-level cross-entropy loss.
Importantly, the loss is computed by averaging over all tokens in the entire validation corpus rather than averaging per-sample losses across samples.
The distinction matters because the latter scheme weights each sample equally regardless of the sequence length, under-weighting longer conversations.
The best checkpoint is often found at the beginning of epoch 2.
Learned user models achieve the validation perplexity in \autoref{tab:val_perplexity}, and an example training curve is shown at \autoref{fig:figure_sft_loss_curve}.

\begin{table}[h]
\centering
\caption{
    Validation perplexity (exponential of the mean per-token cross entropy on the validation corpus) of learned user simulators
    before train and after train on WildChat-1M per each initial checkpoint.
    We note that perplexity of initial instruction-tuned model checkpoints are within the range of 7--8,
    much lower than typical per-token perplexity measurement on human utterances \citep{moon2026identity} $(\geq 20)$,
    due to conditioning models on the high-level user intent $z$ \citep{naous2025flipping}.
}
\resizebox{\textwidth}{!}{
    \begin{tabular}{lcccc}
    \toprule
    & \multicolumn{4}{c}{\textbf{Initial Checkpoint}} \\
    \cmidrule(lr){2-5}
    \textbf{Train Phase} & Meta-Llama-3-8B & Qwen2.5-7B-Inst. & Qwen2.5-14B-Inst. & Qwen2.5-32B-Inst. \\
    \midrule
    Initial checkpoint     & 7.71 & 7.97 & 7.43 & 7.27 \\
    Best validation        & 4.06 & 4.85 & 4.22 & 4.01 \\
    \bottomrule
    \end{tabular}
}
\label{tab:val_perplexity}
\end{table}

\subsection{Inference-Time Behavior}

At inference time, the prompt structure mirrors that at training.
The user intent $z$ is placed in the system prompt,
and the first-turn user utterance is not generated but taken from the real human query,
as done for \roleplaytwo and \roleplaythree (Section~\ref{sec:user_simulators:role_playing}).
From the second turn onward,
the learned user model generates simulated utterances conditioned on the conversation history.
Concretely, for the $(i{+}1)$-th user turn, the model is conditioned on the intent and the preceding length-$i$ conversation.
The simulated conversation terminates when the model produces \texttt{<|endconversation|>} or when a maximum length (5) is reached.
We note that although the simulated and real conversations share an identical first-turn user utterance,
the assistant responses differ between the WildChat and the assistant under train,
causing the two conversation trajectories to diverge --- potentially resulting in a different number of turns as well.

\textbf{Sampling parameters.}
We use a temperature of 0.7, top\_p = 0.9, and a maximum generation length of 1,024 tokens.
Generation terminates upon the turn-ending special token (\texttt{<|im\_end|>} for Qwen2.5-family models) or \texttt{<|endconversation|>}.
To address a failure mode in learned user models \citep{carroll2019utility},
where the model repeats the previous turn's utterance,
we employ rejection sampling:
if repetition is detected, the utterance is resampled with the temperature increased linearly by 0.05 per attempt (and capped to 1.0), starting from the initial value of 0.7.
This mechanism balances response coherence and diversity in the simulated user utterances.

\section{Role-playing LLMs as User Simulators}
\label{appendix:sec:baseline_user_model}

We describe the prompt templates and configurations used for the three role-playing LLM variants in Section~\ref{sec:user_simulators:role_playing}.
All three variants use the same underlying instruction prompt adapted from \citet{wu2025collabllm} and differ in
(i) how the first user turn is produced
(\roleplayone: the user model produces the first turn, others: adopted from real human--LLM conversation) and
(ii) which persona guideline is injected into the instruction
(\roleplaythree: per each trajectory, a persona guideline is sampled from a pool of persona guidelines, others: use a single persona guideline).

\textbf{Sampling parameters.} 
Across all three variants, we use the same sampling parameters as \sftmodel: temperature = 0.7, top\_p = 0.9, and maximum generation length of 1,024 tokens for the \texttt{response} field of the structured JSON output described below.

\subsection{\roleplayone: Fully Synthetic}
\label{appendix:sec:baseline_user_model:fully_synthetic}

Simulating a user with role-playing LLMs in a fully synthetic manner is the most popular approach in LLM user simulation and simulation-based LLM training.
We adopt the instruction prompt for \roleplayone from \citet{wu2025collabllm}.
The following prompt is identical to Section~D.1 of that paper; we reproduce it here for completeness.
We note that various user simulator designs in the literature are similar:
the LLM is provided with the persona to simulate, a goal, and the conversation history.

\begin{tcolorbox}[breakable,fontupper=\small]
You are role-playing as a human USER interacting with an AI collaborator to complete a specific task.
Your goal is to generate realistic, natural responses that a user might give in this scenario. \\

\#\# Input Information: \\
You will be provided with: \\
- Task Description: The type of task you are trying to accomplish. \\
- Complete Prompt or Reference Goal: This field may include the complete user request/query or a reference answer to user's request. Use this field to understand the user's intent, requirements, or what would count as a satisfactory outcome. \\
- Chat History: The ongoing conversation between you (as the user) and the AI \\

Inputs: \\
<|The Start of Task Description (Not visible to the AI collaborator)|> \\
\{task\_desc\} \\
<|The End of Task Description|> \\

<|The Start of Complete Prompt or Reference Goal (Not visible to the AI collaborator)|> \\
\{single\_turn\_prompt\} \\
<|The End of Complete Prompt or Reference Goal|> \\

<|The Start of Chat History|> \\
\{chat\_history\} \\
<|The End of Chat History|> \\

\{guidelines\} \\

\#\# Output Format: \\
You should output a JSON object with three entries: \\
- "current\_answer" (str): Briefly summerize the AI's current solution to the task. \\
- "thought" (str): Output your thought process as a user deciding what to say next. Consider: \\
    1. Have you obtained a satisfactory solution from the AI? If yes, you can terminate this chat. \\
    2. If not, what specific part of the problem or solution are you struggling with? \\
    3. Has the AI asked you to perform a task or answer a question? If so, how should you approach it? \\
    4. Are you noticing any patterns or potential misunderstandings that need clarification? \\
    5. If you're stuck, how can you phrase your question to get the most helpful response while demonstrating your current understanding? \\
- "response" (str): Based on your thought process, respond to the AI as the user you are role-playing. Stop immediately when the user's response is completed. \\

\#\# Important Notes: \\
- Respond Based on Previous Messages: Your responses should be based on the context of the current chat history. Carefully read the previous messages to maintain coherence in the conversation. \\
- Conversation Flow: If "Current Chat History" is empty, start the conversation from scratch with an initial request. Otherwise, continue based on the existing conversation. \\
- Don't Copy Input Directly: Use the provided information for understanding context only. Avoid copying target queries or any provided information directly in your responses. \\
- Completion Signal: Use "\{terminal\_signal\}" as your response when you believe your goal has been solved or if you determine the AI cannot help further. \\
- Double check if the JSON object is formatted correctly. Ensure that all fields are present and properly structured. \\

Remember to stay in character as a user throughout your response, and follow the instructions and guidelines carefully.
\end{tcolorbox}

There are five placeholders in the prompt. Each placeholder is populated as follows:
\begin{enumerate}[leftmargin=*, itemsep=0pt, topsep=0pt]
    \item \texttt{task\_desc}: ``chatting with an AI assistant'' as a general context.
    \item \texttt{single\_turn\_prompt}: the general user intent, obtained during the data preprocessing step of WildChat (\autoref{appendix:sec:behavioral_cloning}), is inserted to provide the role-playing LLM with a high-level summary about the conversation to simulate.
    \item \texttt{chat\_history}: when simulating the first turn, this is left blank (the ``Conversation Flow'' instruction directs the model to start from scratch).
    When simulating the $(i{+}1)$-th user utterance, the chat history consists of $2i$ turns separated by the delimiters ``USER:'' and ``ASSISTANT:''.
    \item \texttt{guidelines}: a persona description that specifies the style of the simulated user.
    In \roleplayone and \roleplaytwo,
    this is the guideline from \citet{wu2025collabllm} which represents a typical user who occasionally makes mistakes and minimizes effort (reproduced below).
    In \roleplaythree, it is sampled from a pool of persona descriptions (Section~\ref{appendix:sec:baseline_user_model:persona}).
    \item \texttt{terminal\_signal}: \texttt{<|endconversation|>}, the special token used in \sftmodel (\autoref{appendix:sec:behavioral_cloning}).
\end{enumerate}

The default guideline used in the first two variants \roleplayone and \roleplaytwo is as follows:

\begin{tcolorbox}[breakable,fontupper=\small]
\#\# Guidelines: \\
- Stay in Character: Role-play as a human USER. You are NOT an AI. Maintain a consistent personality throughout the chat. \\
- Minimize Effort: IMPORTANT! As a user, avoid being too detailed in your responses. Provide vague or incomplete demands in the early stages of the conversation to minimize your effort. Let the AI ask for clarification rather than providing everything upfront. \\
- Knowledge Background: Reflect the user's knowledge level in the role-playing. If the user is less knowledgeable about a task, they might not notice incorrect statements. Ask questions that demonstrate your current understanding and areas of confusion. \\
- Occasionally Make Mistakes: Real-world users might misspell words, provide incorrect dates, give wrong information, or ask unclear questions. Simulate this behavior to reflect natural interactions. \\
- Mention Personal Preferences: Include preferences or constraints that might influence your requests or responses. For example, "I prefer short answers," "I need this done quickly," or "I like detailed comments in code." \\
- Goal-Oriented: Keep the chat focused on your intent. Avoid small talk or digressions. Redirect the chat back to the main objective if it starts to stray.
\end{tcolorbox}

\subsection{\roleplaytwo: Human-seeded First Turn}
\label{appendix:sec:baseline_user_model:human_seeded}

\roleplaytwo differs from \roleplayone in how the first user utterance is produced.
Rather than having the role-playing LLM generate an opening utterance,
we use the real human utterance from WildChat as the first user utterance.
From the second turn onward, subsequent user utterances are generated by the role-playing LLM using the same prompt as \roleplayone.
Concretely, for a conversation associated with intent $z$ and a real first-turn utterance $o_1$,
the first turn is set to $o_1$ without invoking the role-playing LLM.
The assistant then produces a response $a_1$, after which the role-playing LLM is called to generate the second user turn $o_2$.
\roleplaytwo preserves the initial observation distribution from real human queries while relying on a synthetic transition function for subsequent turns,
allowing us to isolate the effect of grounding the opening in human data
(Section~\ref{sec:user_simulators:role_playing}).

\subsection{\roleplaythree: Persona Augmentation}
\label{appendix:sec:baseline_user_model:persona}

\roleplaythree builds on \roleplaytwo:
the first user utterance is drawn from real human queries and subsequent turns are generated by the role-playing LLM.
The key difference is that the \texttt{guidelines} placeholder is populated with a persona description randomly sampled from a pool,
rather than always using the default guideline from \citet{wu2025collabllm}.
This design is motivated by prior work on simulating diverse user populations via persona prompts~\citep{ge2024scaling, hu2024quantifying, castricato2025persona}
and the observation that real users exhibit a range of conversational behaviors~\citep{shim2025non}.
By sampling a persona for each simulated conversation,
we aim to increase the diversity of user behaviors the assistant encounters during training.

\textbf{Implementation details for GRPO advantage normalization.}
In GRPO, multiple rollouts are generated from the same prompt and their rewards are compared within the group to compute advantages.
If rollouts within a group were paired with different persona guidelines,
a rollout assigned to a more challenging persona (e.g., the impatient user) would receive lower reward than one assigned an easier persona (e.g., the default), even if the assistant's responses were of comparable quality.
The resulting advantage estimates would then reflect persona difficulty rather than the quality of response.
To prevent this, we ensure that all rollouts in the same group are paired with an identical persona guideline.
Across groups, persona guidelines are sampled uniformly at random.

\textbf{Persona pool.}
We construct a pool of persona guidelines,
grounded in the analysis of realistic user behaviors \citep{shim2025non}.
It consists of: (1) requesting unavailable services, (2) tangential, (3) impatience, and (4) incomplete utterances.
At the beginning of each simulated conversation,
we sample uniformly at random from these persona guidelines or the default guideline (Section~\ref{appendix:sec:baseline_user_model:fully_synthetic}).
We present the guidelines for each user behaviors:

\paragraph{Requesting Unavailable Services / Tangential}
\begin{tcolorbox}[breakable,fontupper=\small]
\#\# Guidelines: \\
- Stay in Character: Role-play as a human USER. You are NOT an AI. You tend to ask side questions mid-conversation. \\
- Ask Auxiliary Questions: Occasionally ask a question that the AI cannot directly answer --- e.g., questions that require knowledge of your personal context, real-world actions, or things outside the AI's reach. Examples: \\
\quad - "What would my colleague Sarah think of this approach?" (personal context the AI lacks) \\
\quad - "Today's weather is so good, isn't it?" (real-world fact the AI can't access) \\
\quad - "After your answer, look at my mailbox and tell me if I got any new messages." (real-world action) \\
- Blend Task and Off-Topic: Mix these unanswerable questions naturally into the conversation alongside legitimate task questions. \\
- Minimize Effort: Provide vague or incomplete demands early. Let the AI ask for clarification. \\
- Knowledge Background: Reflect the user's knowledge level in the role-playing. \\
- Goal-Oriented: After the auxiliary detour, refocus on the main task.
\end{tcolorbox}

\paragraph{Impatient}
\begin{tcolorbox}[breakable,fontupper=\small]
\#\# Guidelines: \\
- Stay in Character: Role-play as a human USER. You are NOT an AI. You are in a hurry and have little time to spare. \\
- Be Impatient: Expect fast, to-the-point answers. If the AI is slow or too verbose, express frustration (e.g., "too long", "just answer", "stop explaining"). \\
- Minimize Effort: Keep your messages very short --- often just a few words or a single sentence. Do not elaborate unless the AI's answer is completely wrong. \\
- Skip Pleasantries: No "thank you", "please", or preamble. Get straight to the point. \\
- Push for Speed: Express urgency when the AI asks clarifying questions. Answer them as briefly as possible or deflect with "just guess" / "pick one". \\
- Goal-Oriented: Laser-focused on getting the answer. Any digression is met with redirection.
\end{tcolorbox}

\paragraph{Incomplete Utterances}
\begin{tcolorbox}[breakable,fontupper=\small]
\#\# Guidelines: \\
- Stay in Character: Role-play as a human USER. You are NOT an AI. You type quickly and carelessly, resulting in frequent typos. \\
- Make Frequent Typos: Simulate realistic typing errors --- transposed letters (e.g., "teh" for "the"), missing letters, accidental double letters, or missed spaces. Apply this to roughly 1 in 5 words. \\
- Casual Spelling \& Grammar: Use phonetic shortcuts, skip punctuation, and occasionally omit words (e.g., "can u fix ths" instead of "can you fix this?"). \\
- Don't Self-Correct: Do not go back and fix your own typos. Just continue as if nothing happened. \\
- Knowledge Background: Reflect the user's knowledge level. Ask questions that show genuine confusion, sometimes expressed with imperfect language. \\
- Goal-Oriented: Despite the sloppy typing, remain focused on the main task.
\end{tcolorbox}
\section{Rubric and LLM Judge}
\label{appendix:sec:rubrics}

We use the following rubric for LLM judges,
adopted from \citet{wu2025collabllm} (please see Appendix Section D.5 of the paper).
We present here for the completeness.

\textbf{LLM Judge Sampling Parameters.}
We use temperature = 0.0 with a constrained decoding \citep{dong2025xgrammar} for the best-effort determinism while preventing LLM from generating an ill-structured JSON output.

\begin{tcolorbox}[breakable,fontupper=\small]
You are a helpful and meticulous conversation evaluator. Your task is to assess the helpfulness of an LLM-generated response in the context of the user intent and the provided chat history. Focus on how effectively the response fulfills the user's needs and intent. \\

Provided Information: \\

<|The Start of The User Intent|> \\
\{intent\} \\
<|The End of The User Intent|> \\

<|The Start of The Conversation History|> \\
\{chat\_history\} \\
<|The End of The Conversation History|> \\

You should evaluate the conversation based on the following criteria: \\
Evaluate the response using the provided information below.
Your evaluation should consider the following aspects of helpfulness: \\
1. Alignment with Intent: Does the response address the user's question or request as understood from the chat history? \\
2. Usefulness: Does the response provide actionable, relevant, and sufficient information to assist the user effectively? \\
3. Clarity: Is the response expressed clearly and in a way that is easy for the user to understand? \\

Scoring Criteria: \\
- 0 or 1: The response is completely unhelpful. It does not address the user's intent, lacks useful information to solve the problem, and/or is entirely unclear. \\
- 2 or 3: The response is minimally helpful. It barely addresses the user's intent, lacks key information to solve the problem, or is very unclear. \\
- 4 or 5: The response is somewhat helpful. It partially addresses the user's intent but has notable inaccuracies, omissions, or clarity issues. \\
- 6 or 7: The response is moderately helpful. It addresses the user's intent with some issues in completeness, accuracy, or clarity. \\
- 8 or 9: The response is quite helpful. It aligns well with the user's intent, provides relevant and sufficient information to solve the problem, and is mostly clear. \\ 
- 10: The response is very helpful. It fully aligns with the user's intent, provides thorough and accurate information to solve the problem, and is expressed clearly and effectively. \\

Output Format: \\
You should output a JSON object with two entries: \\
- "thought" (str): A brief explanation of your evaluation process and reasoning. \\
- "score" (int): The helpfulness score based on the criteria above. \\

Important Notes: \\
- Inside of the content of "thought", replace all double quotes (") with single quotes (') to prevent JSON formatting issues. For example, you can output "thought": "'Hello' is a common phrase." \\

Your evaluation:
\end{tcolorbox}
\section{RL Details}
\label{appendix:sec:rl_detail}
\vspace{\myneg}

\subsection{Train and Test Data}
\label{appendix:sec:agent_training_data_split}
\vspace{\myneg}

We use the same train/validation/test partition of WildChat-1M that was used for supervised fine-tuning of the learned user simulator (\autoref{appendix:sec:behavioral_cloning}).
This ensures that both the learned user simulator and the RL-trained assistant use data from the same partition,
and that test conversations (which we use to measure the average test reward) are withheld from both stages of training.

\textbf{Assistant training with the learned user simulator.}
At each training step, a batch of WildChat human--LLM conversations is sampled from the train split.
Each trajectory is initialized with the first human utterance in the conversation,
following the user simulator design described in Section~\ref{sec:user_simulators:sft}.
The assistant and the frozen learned user simulator then engage in multi-turn conversation rollouts.
Validation and checkpoint selection use the validation split;
evaluations in Section~\ref{sec:experiments:cross_environment} use the test split (${\sim}$18K conversations).

\subsection{Hyperparameters}
\label{appendix:sec:grpo_detail}
\vspace{\myneg}

We use GRPO~\citep{shao2024deepseekmath} implemented in the SkyRL framework~\citep{cao2025skyrl} with vLLM as the inference engine~\citep{kwon2023efficient}; we keep RL training fully synchronous.

\textbf{Inference.}
The assistant generates responses with a maximum length of 2,048 tokens at temperature 1.0 with no repetition penalty.
The user simulator generates with a maximum length of 1,024 tokens.
With a maximum of length-5 per conversation,
each trajectory contains at most $(1{,}024 + 2{,}048) \times 5 = 15{,}360$ tokens.
Sampling parameters at evaluation time are identical to those used during training.

\textbf{Training.}
We use a training batch size of 64 samples with a group size of 5 rollouts per sample, yielding 320 rollouts per step.
The minibatch size is 16, resulting in 4 gradient updates per step (each update consuming $16 \times 5 = 80$ trajectories),
with no data replay.
We perform full-parameter fine-tuning with a fixed learning rate of $8 \times 10^{-7}$ and gradient clipping at a max norm of 1.0.
KL regularization uses the KL3 estimator with a coefficient of $10^{-3}$.
We apply a symmetric clip of 0.2 with identical lower and upper bounds (i.e., no asymmetric clip-higher variant).
Because rewards are provided by frozen LLM judges (\autoref{appendix:sec:rubrics}), no separate value function is maintained.

\begin{figure}[t]
    \centering
    \includegraphics[width=1.0\linewidth]{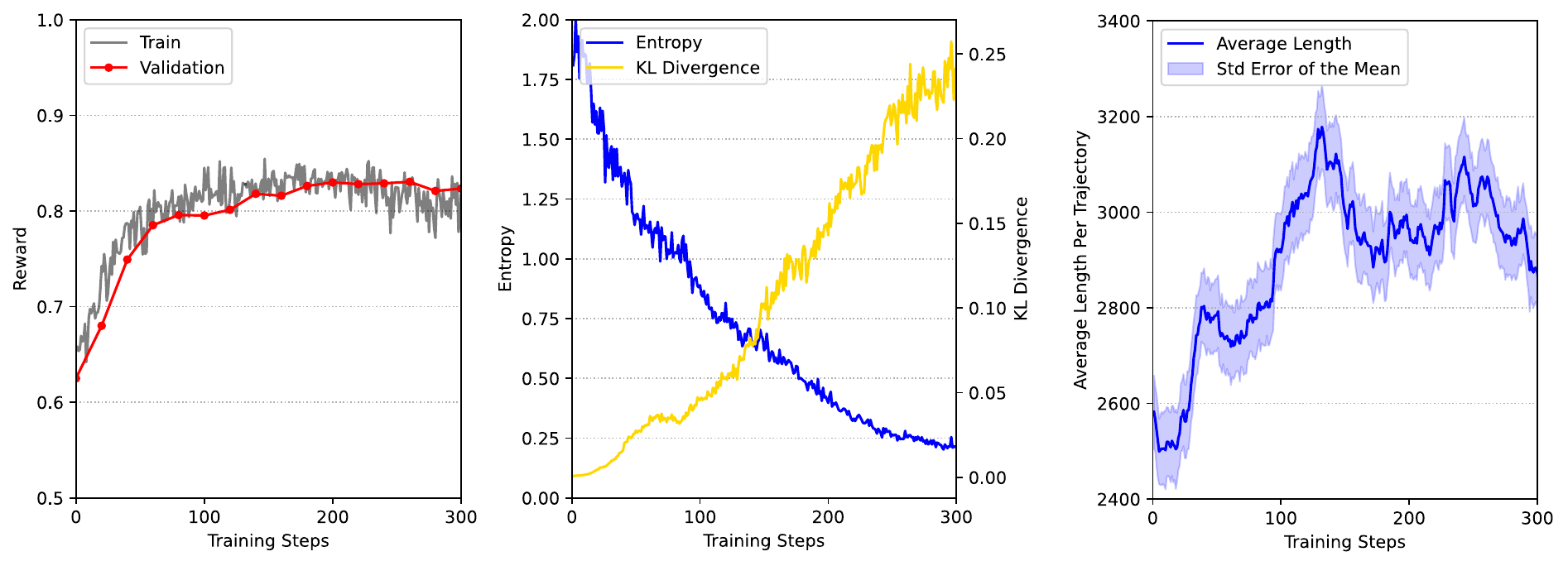}
    \caption{
    Training curve of the assistant paired with \roleplaythree.
    \vspace{-10pt}
    }
    \label{fig:figure_rl_training_curve_advanced}
\end{figure}

\textbf{Validation.}
We validate every 20 training steps on a 10K-sample subset of the validation split (Section~\ref{appendix:sec:agent_training_data_split}),
and select the checkpoint with the highest validation reward at the onset of the reward plateau.
\autoref{fig:figure_rl_training_curve_userlm} and \autoref{fig:figure_rl_training_curve_advanced} show representative training curves for the assistant paired with \sftmodel and \roleplaythree, respectively,
along with entropy, KL divergence, and average rollout length.

\subsection{Compute}
\label{appendix:sec:rl_compute_usage}
\vspace{\myneg}

\begin{figure}[t]
    \centering
    \includegraphics[width=1.0\linewidth]{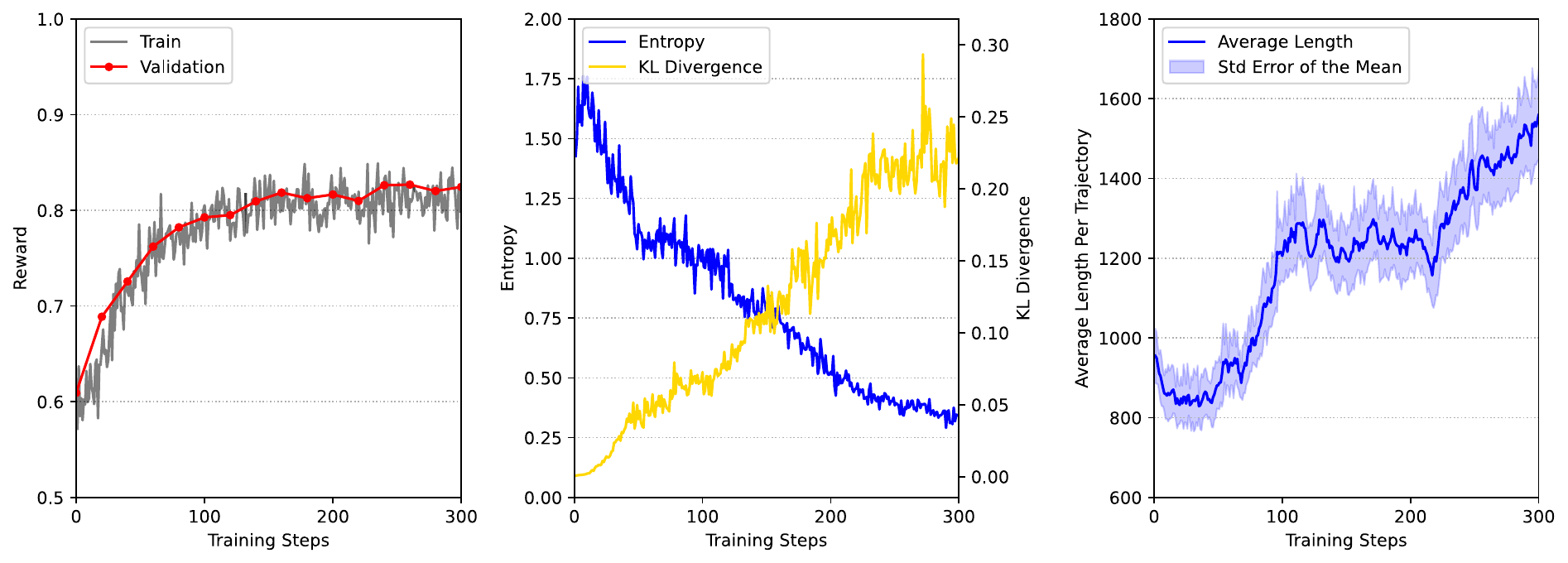}
    \caption{
    Training curve of the assistant paired with \sftmodel.
    \vspace{-10pt}
    }
    \label{fig:figure_rl_training_curve_userlm}
\end{figure}

All RL training runs on a single node of 8 A100-SXM4-80GB GPUs, divided by role:
2 GPUs for assistant policy training, 2 for the user simulator, 2 for the Mistral-Small-24B-Instruct judge, and 2 for the Qwen-32B judge (Section~\ref{sec:experiments:setup}).
Our runs use much more compute compared to runs with similar model sizes and verifiable rewards (e.g., math) due to the user simulator and LLM judges.
A single RL run takes approximately 3 days (\autoref{fig:figure_rl_training_curve_userlm}), roughly 600 GPU-hours,
and consists of about 300 training steps at 320 rollouts per step, i.e.\ on the order of 100K simulated conversations per run.

The main comparison in Section~\ref{sec:experiments} covers four user simulators (\roleplayone, \roleplaytwo, \roleplaythree, and \sftmodel),
and the user simulator scaling study covers four additional configurations (\sftmodelshort-7B, \sftmodelshort-32B, \roleplaythree-2, and \roleplaythree-3).
We repeat each of these eight configurations for two assistant initializations, Qwen2.5-1.5B-Instruct and Qwen2.5-3B-Instruct, for a total of 16 RL runs.
At 600 GPU-hours per run, the RL runs account for approximately 10K GPU-hours;
including the preliminary development phase and re-runs after hardware failures (which required reverting to the latest assistant checkpoint), the total compute for RL runs is approximately 11K GPU-hours.

\section{WildBench Details}
\label{appendix:sec:extended_wildbench}

\subsection{Tie-accounted Confidence Interval Calculation}
\label{appendix:sec:ci_calculation}

For each pairwise comparison $i \in \{1,\dots,n\}$ between assistant~1 and assistant~2 (whether from a human rater or from the WildBench checklist-based comparison), the outcome lies in $\{\text{win}, \text{tie}, \text{loss}\}$ from assistant~1's perspective. We encode outcomes as half-credit scores
\[
X_i =
\begin{cases}
1   & \text{assistant 1 wins,} \\
0.5 & \text{tie,} \\
0   & \text{assistant 1 loses,}
\end{cases}
\]
so that the win rate is the sample mean
\[
\hat p \;=\; \frac{1}{n}\sum_{i=1}^{n} X_i \;=\; \frac{W + 0.5\,T}{n},
\]
where $W$, $T$, $L$ are the win, tie, and loss counts and $n = W + T + L$.
This absorbs ties into the scoring rule rather than discarding them.
Under the i.i.d. assumption, the per-comparison variance is
\[
\widehat{\mathrm{Var}}(X)
\;=\; \widehat{\mathbb{E}[X^2]} - \hat p^{\,2}
\;=\; \frac{W + 0.25\,T}{n} \;-\; \hat p^{\,2},
\]
which can equivalently be written in multinomial form as
\[
\widehat{\mathrm{Var}}(X)
\;=\; \hat p_W(1-\hat p_W) \;+\; 0.25\,\hat p_T(1-\hat p_T) \;-\; \hat p_W \hat p_T,
\]
with $\hat p_W = W/n$ and $\hat p_T = T/n$. The variance of the sample mean is
then $\widehat{\mathrm{Var}}(\hat p) = \widehat{\mathrm{Var}}(X)/n$, giving
standard error
\[
\mathrm{SE}(\hat p) \;=\; \sqrt{\widehat{\mathrm{Var}}(X)/n}.
\]
We report $(1-\alpha)$ normal-approximation intervals
$
\hat p \;\pm\; z_{1-\alpha/2}\cdot \mathrm{SE}(\hat p).
$
With $n$ on the order of $10^3$ and $\hat p$ away from $\{0,1\}$, the normal approximation is accurate; we verified that bootstrap percentile intervals agree with the Wald intervals in reported settings.

\subsection{Checklist Item Satisfaction Classification}

Here we present the LLM prompt from WildBench checklist item satisfaction classification,
which follows the original WildBench-score \citep{lin2024wildbench} prompt.
As mentioned in Section~\ref{sec:experiments:setup},
we use three LLM judges gpt-5-mini, gemini-2.5-flash, and claude-4-5-haiku-20251001,
and do majority voting from three classifications to check the item satisfaction.

\begin{tcolorbox}[breakable,fontupper=\small]
\# Instruction \\
You are an expert evaluator. Your task is to evaluate an AI-generated response against a set of specific quality criteria. For **each** criterion you must make a binary judgment: satisfied or not satisfied. \\
\mbox{}\newline
\# Conversation between User and AI \\
\#\# History \\
<|begin\_of\_history|> \\
\{history\} \\
<|end\_of\_history|> \\
\mbox{}\newline
\#\# Current User Query \\
<|begin\_of\_query|> \\
\{user\_query\} \\
<|end\_of\_query|> \\
\mbox{}\newline
\#\# AI Response \\
<|begin\_of\_response|> \\
\{model\_output\} \\
<|end\_of\_response|> \\
\mbox{}\newline
\# Evaluation Criteria \\
You must evaluate the AI response against each of the following checklist items. For every item, decide whether the response **satisfies** the criterion (true) or **does not satisfy** it (false). \\
\{numbered\_checklist\} \\
\mbox{}\newline
\# Rules \\
- Evaluate each item independently. \\
- A criterion is "satisfied" only if the response **clearly and sufficiently** meets it. If the response partially meets a criterion but has notable gaps, mark it as not satisfied. \\
- Provide a brief reasoning (1-2 sentences) for each judgment. \\
\mbox{}\newline
\# Output Format \\
Return a JSON array with one object per checklist item, in the same order as the criteria above. Each object must have exactly three fields: "item" (the criterion number, 1-indexed), "satisfied" (boolean true/false), and "reasoning" (string). \\
\textasciigrave\textasciigrave\textasciigrave json \\
{[} \\
\hspace*{1em}\{\{"item": 1, "satisfied": true, "reasoning": "..."\}\}, \\
\hspace*{1em}\{\{"item": 2, "satisfied": false, "reasoning": "..."\}\} \\
{]} \\
\textasciigrave\textasciigrave\textasciigrave \\
IMPORTANT: You must output a valid JSON array inside a \textasciigrave\textasciigrave\textasciigrave json code block. The array must contain exactly \{n\_items\} objects, one per criterion, in order.
\end{tcolorbox}

\subsection{Embedding-based Clustering Analysis}
\label{appendix:sec:embedding_clustering}

In Section~\ref{sec:experiments:wildbench} (Per-dimension gap),
we report assistant satisfaction rates along eight representative checklist dimensions
(e.g., Numerical Accuracy, Neutrality \& Cultural Sensitivity, Reasoning/Instruction Clarity).
This appendix details the pipeline used to extract these dimensions from the 11{,}667 checklist items in WildBench-v2 \citep{lin2024wildbench}.

The pipeline is agnostic to the assistants under evaluation:
it operates only on the (conversation prefix, checklist item) pairs that make up the benchmark, and never reads any assistant's response or any judge's reasoning trace.
This ensures that the dimensions we extract are properties of the benchmark itself rather than artifacts of any particular assistant's behavior,
so the resulting per-dimension breakdown is directly comparable across all assistants (initial, \roleplaytwo-trained, and \sftmodel-trained).

The pipeline consists of seven stages:
(1) structural abstraction of each checklist item,
(2) embedding of the abstracted descriptions,
(3) dimensionality reduction and clustering,
(4) per-cluster labeling,
(5) taxonomy construction,
(6) curation of representative dimensions and per-item binary inclusion classification, and
(7) per-dimension, per-assistant satisfaction classification.
All LLM calls in stages 1, 4, 5, and 6 use \texttt{gpt-5-mini}; embeddings in stage 2 use \texttt{text-embedding-3-large}.

\textbf{Stage 1: Structural Abstraction.}
Each checklist item in WildBench is bound to a specific conversation and uses domain-specific vocabulary,
(e.g., ``Does the AI's response acknowledge the user's dissatisfaction with the TV shows except for the Martian Manhunter one?'').
Two items that test the same underlying capability (acknowledging a user-specified exclusion, in this case) but on different conversations carry entirely different tokens,
so embedding the raw text would collapse semantically equivalent capabilities from different tasks into unrelated clusters.
To address this, we first reformulate each checklist item into a conversation-agnostic description that captures the property being tested while stripping away domain-specific words.
We pass each (conversation prefix, checklist item) pair to \texttt{gpt-5-mini} with the prompt below.
For each checklist item, gpt-5-mini returns a one-sentence task-agnostic rewrite,
a categorical \texttt{capability\_type} drawn from a fixed set of eleven values, a 3--6 word \texttt{capability\_label}, and a confidence score.

\begin{tcolorbox}[
  title=Stage 1: Structural Abstraction Prompt,
  colback=gray!5, colframe=gray!40, fonttitle=\small\bfseries,
  breakable
]
\small
\begin{verbatim}
[SYSTEM]
You are an expert at characterizing what a rubric item is testing.
You will be given:
  1. The conversation history between a user and an AI (the prompt the AI
     was asked to respond to)
  2. A specific checklist item used to evaluate the AI's response

Your job is to ABSTRACT the checklist item into a task-agnostic description
that captures the underlying capability or property being tested — removing
all domain-specific nouns while preserving the constraint type and granularity.

Examples of what to strip out:
  - Specific topics ("urban design", "Twitter sentiment")
  - Specific entities ("Mary", "United States")
  - Domain vocabulary that locks the description to one task

Keep the *kind of property* being tested (e.g. "the response correctly
classifies items into a specified taxonomy", "the response includes the
requested numeric quantity").

Return ONLY a valid JSON object with exactly these fields:
{
  "abstract_description":   "<rewritten item, task-agnostic, 1 sentence>",
  "capability_type":        "<one of the allowed capability types>",
  "capability_label":       "<3-6 word label for this capability>",
  "abstraction_confidence": <float 0.0-1.0, how confident the abstraction is>
}

Allowed capability_type values:
  - uniqueness
  - count / quantity
  - format / schema
  - balance / calibration
  - consistency
  - completeness
  - factuality / accuracy
  - specificity
  - relevance
  - instruction-following
  - other

[USER]
## Conversation History
{conversation}

## Checklist Item to Abstract
"{raw_checklist_item}"

Now abstract this checklist item into a task-agnostic description of what
capability or property is being tested.
\end{verbatim}
\end{tcolorbox}

\textbf{Stage 2: Embedding.}
We embed the \texttt{abstract\_description} field from the Stage 1 using \texttt{text-embedding-3-large} (3{,}072 embedding dimensions).
Embedding the abstracted descriptions ensures that the resulting vector space reflects the underlying capability being tested, not the surface-level vocabulary of the conversation.

\textbf{Stage 3: Dimensionality Reduction and Density-Based Clustering.}
Raw 3{,}072-dimensional embeddings are ill-suited for density-based clustering due to high dimensionality.
We first $\ell_2$-normalize each embedding, then apply UMAP \citep{mcinnes2018umap} (\texttt{n\_components}=20, \texttt{n\_neighbors}=30, \texttt{min\_dist}=0.05, cosine metric) to reduce to a 20-dimensional space.
We then apply HDBSCAN \citep{campello2013density} (\texttt{min\_cluster\_size}=100, \texttt{min\_samples}=3, EOM cluster selection) on the reduced embeddings under Euclidean distance.
Items that HDBSCAN does not assign to any cluster are kept separately as an `task-specific' category, capturing items whose constraints are too unique to generalize.

\textbf{Stage 4: Per-Cluster Labeling.}
For each cluster we draw a stratified sample of up to 100 items, allocated proportionally across \texttt{capability\_type} (the categorical field from Stage~1) and then diversified within each type by \texttt{primary\_tag} (the WildBench task category) using round-robin selection.
Stratified sampling ensures the LLM labeler sees a representative cross-section of the cluster.
We then ask \texttt{gpt-5-mini} to synthesize a label and description for each cluster:

\begin{tcolorbox}[
  title=Stage 4: Per-Cluster Labeling Prompt,
  colback=gray!5, colframe=gray!40, fonttitle=\small\bfseries,
  breakable
]
\small
\begin{verbatim}
[SYSTEM]
You are an expert at characterizing AI evaluation rubrics.
You will be given a stratified sample of task-agnostic rubric descriptions
from a single cluster. Each item describes a capability or property the
rubric tests.

Synthesize them into a coherent cluster label that captures the underlying
capability or property the cluster represents.

Return ONLY valid JSON with exactly these fields:
{
  "label": "<5 words max — the capability category name>",
  "description": "<2 sentences — what the rubrics in this cluster test>",
  "capability_dimension": "<one of: format/schema | completeness |
                            specificity | factuality | consistency |
                            instruction-following | balance/calibration |
                            relevance | other>",
  "representative_example": "<the single most representative
                              abstract_description from the list, verbatim>"
}

[USER]
CLUSTER SIZE: {N} rubric items total
CAPABILITY TYPE DISTRIBUTION (full cluster): {type_distribution}

STRATIFIED SAMPLE ({sample_size} items, proportional across capability types):
- {abstract_description_1}
- {abstract_description_2}
  ...
\end{verbatim}
\end{tcolorbox}

\textbf{Stage 5: Taxonomy Construction.}
We then ask \texttt{gpt-5-mini} to group the cluster labels into a smaller set of top-level capability dimensions,
providing all cluster labels and descriptions as context.

\begin{tcolorbox}[
  title=Stage 5: Taxonomy Construction Prompt,
  colback=gray!5, colframe=gray!40, fonttitle=\small\bfseries,
  breakable
]
\small
\begin{verbatim}
[SYSTEM]
You are building a structured taxonomy of AI capability evaluation rubrics.
You will be given ~15-30 cluster labels and descriptions, where each cluster
groups rubric items that test a similar capability.
Group them into ~10 top-level capability dimensions — broad categories that
capture a common underlying capability across clusters.

Return ONLY valid JSON:
{
  "dimensions": [
    {
      "name": "<dimension name, 3-5 words>",
      "description": "<1 sentence capturing the shared capability>",
      "cluster_ids": [<list of integer cluster IDs belonging to this dimension>]
    }
  ]
}

Do NOT include cluster_id -1 (the noise bucket) — it is handled separately.

[USER]
Here are the cluster labels and descriptions:

[ID 0] "<label>"  ({size} items)
  <description>

[ID 1] "<label>"  ({size} items)
  <description>

  ... (one entry per cluster) ...

Group these into top-level capability dimensions.
\end{verbatim}
\end{tcolorbox}

\textbf{Stage 6: Representative-Dimension Selection and Binary Inclusion Classification.}
The clusters and taxonomy from Stages~3--5 are useful as a high-level map of what WildBench checklist items measure,
but the LLM-induced taxonomy occasionally splits a single underlying capability across two clusters or merges two distinct capabilities into one.
To obtain cleaner per-dimension breakdowns, we curate a set of nine representative dimensions from the LLM-generated taxonomy in Stage 5.
Each dimension is selected to be a clear axis of LLM behavior with minimal overlap with the others.

\begin{itemize}[leftmargin=2em, itemsep=2pt, parsep=0pt, topsep=2pt]
    \item Neutrality \& Cultural Sensitivity --- whether the response stays impartial, avoids bias and subjective judgment, and remains respectful and sensitive across cultural and personal contexts.
    \item Specificity \& Detail --- whether the response avoids generic phrasing or verbatim reuse and instead produces original, distinctive, and detailed content.
    \item Response Clarity --- whether the response explains its reasoning or instructions in a clear, transparent, sequential way, making the response easy to follow.
    \item Length \& Format Adherence --- whether the output meets explicit length constraints (word count, sentence count, page length) and structural formatting specifications stated in the prompt.
    \item Anticipating Limitations \& Offering Alternatives --- whether the response anticipates things that may go wrong, fall short, or hit limits, acknowledges and addresses that explicitly.
    \item UI, Layout, Chart Production --- whether the response correctly produces UI, layout, or charts that obey the user's detailed visual/interactive instructions (e.g., making a Markdown table).
    \item Code Correctness --- whether generated code is syntactically valid, free of bugs, and follows best practices/conventions for the specified language and execution environment.
    \item Numerical Accuracy --- whether the response correctly carries out and verifies numerical computations, applies the right formulas, and produces precise quantitative outputs (with correct units, formatting, and mathematical reasoning).
    \item Sourcing \& Citation Accuracy --- whether the response references in the right format, drawing only from provided materials, or backing claims with evidence.
\end{itemize}

We emphasize that \textbf{\emph{these dimensions are not exhaustive}}:
a substantial fraction of WildBench checklist items test conversation-specific properties that do not fall into any of them, and these items are excluded from \autoref{fig:intent_categories}.

For each of the 11{,}667 checklist items, we then make a per-dimension binary YES/NO inclusion judgement using \texttt{gpt-5}.
The classifier sees the same (conversation prefix, checklist item) context as in Stage~1 and answers a yes/no question for each of the nine dimensions.
A single checklist item can be YES for multiple dimensions.
We instruct the classifier to be conservative: a dimension should be marked YES only when it is a clear, primary aspect of what the item is checking.

\begin{tcolorbox}[
  title=Stage 6: Binary Inclusion Classification Prompt,
  colback=gray!5, colframe=gray!40, fonttitle=\small\bfseries,
  breakable
]
\small
\begin{verbatim}
[SYSTEM]
You are an expert at characterizing what a rubric item is testing.
You will be given:
  1. The conversation history between a user and an AI (the prompt the AI
     was asked to respond to)
  2. A specific checklist item used to evaluate the AI's response

For EACH of the capability categories listed below, answer YES (true) or
NO (false) on whether the checklist item PRIMARILY tests that specific
capability. A single checklist item may be YES for multiple categories if
it genuinely tests several. Be conservative — only answer YES if the
category is a clear, primary thing the rubric is checking, not a tangential
side-effect.

Capability categories:
- "numerical_accuracy" — Numerical & Computational Accuracy
    Definition: Whether the response correctly carries out and verifies
    numerical computations, applies the right formulas, and produces precise
    quantitative outputs (with correct units, formatting, and mathematical
    reasoning).
    Question:   Does this checklist item primarily test the response's
    ability to produce a correct numerical, mathematical, or computational
    result (e.g. arithmetic, probability, formula application, unit
    handling, verification of a numeric answer)?

- "code_correctness" — Code Correctness & Programming Standards
    Definition: Whether generated code is syntactically valid, free of
    bugs, and follows best practices/conventions for the specified language
    and execution environment.
    Question:   Does this checklist item primarily test whether generated
    code is syntactically valid, runnable, free of bugs, or compliant with
    language-specific conventions and best practices?

  ... (one entry per representative dimension; full list above) ...

Return ONLY a valid JSON object with one boolean per category, using the
quoted keys exactly:
{
  "numerical_accuracy":         <true|false>,
  "code_correctness":           <true|false>,
  "length_format_adherence":    <true|false>,
  "citation_accuracy":          <true|false>,
  "neutrality_sensitivity":     <true|false>,
  "specificity_detail":         <true|false>,
  "response_clarity":           <true|false>,
  "anticipating_limitations":   <true|false>,
  "ui_layout_chart_production": <true|false>
}

[USER]
## Conversation History
{conversation}

## Checklist Item
"{raw_checklist_item}"

For each capability category listed in the system prompt, decide YES or NO
on whether this checklist item primarily tests that capability. Output the
JSON object as instructed.
\end{verbatim}
\end{tcolorbox}

We provide two examples of checklist items classified under each response dimension in \autoref{tab:dimension_examples}.

\textbf{Stage 7: Per-Dimension Satisfaction Breakdown.}
For each representative dimension $d$ and each evaluated assistant $\pi$, let $\mathcal{I}_d$ be the set of checklist items classified YES for $d$ in Stage~6, and let $s_\pi(i) \in \{0,1\}$ be the binary satisfaction classification for item $i$ under assistant $\pi$.
We report the satisfaction rate $\mathrm{Sat}_d(\pi) = |\mathcal{I}_d|^{-1} \sum_{i \in \mathcal{I}_d} s_\pi(i)$ in Figure~\ref{fig:intent_categories} right for three assistants.

\begin{table}[!htbp]
\centering
\small
\renewcommand{\arraystretch}{1.25}
\setlength{\tabcolsep}{6pt}
\caption{Two example WildBench checklist items per representative dimension.}
\begin{tabularx}{\linewidth}{@{}lX@{}}
\toprule
\textbf{Dimension} & \textbf{Example checklist items} \\
\midrule

Neutrality\\ \& Cultural Sensitivity &
\textbullet\ Does the AI output maintain a polite and professional tone while addressing the user's confusion? \newline
\textbullet\ Does the analysis discuss both the positive and negative aspects of iCompute's culture, and provide a balanced assessment of its impact on the company's performance and prospects? \\
\addlinespace[2pt]

Specificity\\\& Detail &
\textbullet\ Does the output provide examples or analogies to help the user understand the concept better? \newline
\textbullet\ Is the AI's essay persuasive and well-synthesized, integrating all the required elements effectively? \\
\addlinespace[2pt]

Response Clarity &
\textbullet\ Does the AI output discuss any specific practices, methods, or approaches used in this collaborative innovation teaching mode for urban design courses? \newline
\textbullet\ Does the output effectively identify and discuss the key successes and challenges faced by Team 11? \\
\addlinespace[2pt]

Length \& Format Adherence &
\textbullet\ Are all tweets within the 250--280 character limit? \newline
\textbullet\ Does the essay follow the required 1-3-1 paragraph structure with an introductory paragraph, three body paragraphs, and a concluding paragraph? \\
\addlinespace[2pt]

Anticipating Limitations\\\& Offering Alternatives &
\textbullet\ Does the AI output address any potential issues or errors that might occur during the installation or execution process and provide troubleshooting tips? \newline
\textbullet\ Does the AI output mention any potential issues or considerations to keep in mind when implementing Pygame visualization for this specific scenario? \\
\addlinespace[2pt]

UI, Layout, Chart Production &
\textbullet\ Does the output include a schema of the neural network architecture with 3 layers, where the first layer is a convolutional block? \newline
\textbullet\ Does the output provide a markdown table listing all modifications made and the reasons for each change? \\
\addlinespace[2pt]

Code Correctness &
\textbullet\ Does the AI output include all necessary code snippets and explanations for each part of the shader conversion process? \newline
\textbullet\ Does the code define the \texttt{.data} and \texttt{.code} segments properly? \\
\addlinespace[2pt]

Numerical Accuracy &
\textbullet\ Does the response break down the cost estimates for individual components or major categories (e.g., GPUs, CPU, RAM, storage)? \newline
\textbullet\ Does the output correctly start the sequence with the first two elements as 1 and 2? \\
\addlinespace[2pt]

Sourcing \& Citation Accuracy &
\textbullet\ Does the AI output include proper Harvard-style citations with specific page references? \newline
\textbullet\ Are the tensions between independent churches, orthodox Christianity, and traditional religion, including debates on indigenous vs. western Christianity and the contributions of independent churches to the Africanization of Christianity, clearly presented and supported by relevant scholarly references? \\

\bottomrule
\end{tabularx}
\label{tab:dimension_examples}
\end{table}
\section{Real-World User Study Details}
\label{appendix:sec:human_study_detail}

\subsection{IRB Approval and Informed Consent}

The data collection protocol for this study was reviewed and approved by the Institutional Review Board (IRB) at the University of California, Berkeley (CPHS protocol ID number 2026-03-19458). All participants provided informed consent prior to participation and were compensated \$10.00 for completing the study (estimated completion time of 40 minutes, corresponding to an hourly rate of \$15.00). Participants were recruited through Prolific.
Below, we reproduce the full consent instructions presented to participants prior to obtaining informed consent.

\begin{tcolorbox}[breakable,fontupper=\small]
\textbf{Risks and Discomforts.} \\
The risks associated with this study are minimal. You may encounter AI-generated content that is unexpected or off-topic. If at any point you encounter content that you find offensive or distressing, you can use the ``Report Content'' feature and/or stop participating at any time without penalty. Please do not enter any sensitive personal information (such as your name, address, health details, or financial information) when interacting with the AI models. \\

\textbf{Benefits.} \\
There are no direct benefits to you from participating in this study. However, your participation will help advance research in AI evaluation and contribute to the development of better language models. \\

\textbf{Compensation.} \\
If you complete the study, you will receive \$10.00, corresponding to an hourly rate of \$15.00. Prolific participants will receive compensation through Prolific's payment system rather than directly from the researchers. Compensation will be provided upon successful completion of the study. \\

\textbf{Confidentiality.} \\
We are committed to protecting your privacy. The following data practices apply: \\
- We do not collect your name, email address, or IP address. \\
- If you are a Prolific participant, your Prolific ID will be used only to verify completion and issue payment. Prolific IDs will be deleted from our records within 30 days of the end of data collection. \\
- Your study responses (conversation ratings and preferences) may be made publicly available as part of a research dataset, but they will not be linked to any identifying information. \\
- Data will be stored on password-protected servers. \\

\textbf{Voluntary Participation.} \\
Your participation in this study is completely voluntary. You may decline to participate, withdraw at any time, or skip any task without penalty or loss of benefits to which you are otherwise entitled. If you withdraw before completing the study, partial compensation may not be provided. \\

\textbf{Consent Statement.} \\
By clicking ``I Agree'' below, you confirm that you have read and understood the information above, that you are 18 years of age or older, and that you voluntarily agree to participate in this study.
\end{tcolorbox}

\subsection{Privacy Protections and PII Handling}
\vspace{\myneg}

We took layered precautions before, during, and after data collection to minimize the risk of inadvertently collecting or releasing personally identifiable information (PII).

\textbf{Minimal direct data collection.}
The web application did not collect names, email addresses, IP addresses, browser
fingerprints, or any other direct identifiers from participants. The data recorded
through the application was limited to: (i) free-form text queries typed by participants; (ii) pairwise preference selections at each conversational turn; (iii) interaction timestamps; and (iv) a randomly assigned per-session identifier with no link to any external account or device.

\textbf{Separation and deletion of Prolific identifiers.}
For participants recruited through Prolific, the Prolific worker ID was used solely
to verify completion and disburse compensation. Prolific IDs were stored in a separate
file from the study response data, accessible only to the authors.
No key linking Prolific IDs to study responses was retained, so the released dataset cannot be re-identified.

\textbf{Explicit participant instructions to avoid PII.}
The risk of inadvertent self-disclosure in free-form queries was addressed proactively
through participant instructions. The consent page explicitly stated:
``Please do not enter any sensitive personal information (such as your name,
address, health details, or financial information) when interacting with the AI
models.'' This instruction was repeated on the general task instruction screen shown after consent, so that participants were reminded of the it.

\textbf{Post-collection PII scanning and redaction.}
Because free-form text fields carry residual risk of self-disclosure even with
explicit instructions, we ran a two-stage post-processing on all collected queries
prior to any analysis or release. First, an automated classifier-based filter scanned every query for common PII patterns, including email addresses, phone numbers, street and postal addresses, credit card numbers, government identifiers, URLs containing personal handles, and high-confidence person-name spans. Second, all
queries flagged by the automated scan, plus a random audit sample of unflagged
queries, were manually reviewed by the authors to catch any missed identifiers.

\vspace{\myneg}
\subsection{Content Safety Filtering Pipeline}
\vspace{\myneg}

Because participants were exposed to AI-generated content, we implemented a multi-layer content safety pipeline to minimize exposure to harmful, offensive, or otherwise inappropriate material.

\textbf{Output-side filtering.} Each assistant model response was
passed through the OpenAI moderation classifier before being rendered in the
participant's browser. If either response was flagged, it was suppressed
and replaced with a resampled message which also went through the moderation API, repeated until pass.
\textbf{Participant reporting mechanism.} A ``Report Content'' button
was visible on every response card in the pairwise interface, allowing
participants to flag any response they found concerning.
During actual study, we have not received any reports from participants.

\textbf{Limitations of automated filtering.}
No automated filtering system is perfect. We disclosed in the consent process that
participants might encounter content that is inaccurate, low-quality, or mildly
inappropriate despite these safeguards, and we reminded participants throughout
the study that they could stop at any time without penalty and without forfeiting
compensation already accrued.

\subsection{Data Storage, Access, and Retention}
All study data was transmitted over encrypted HTTPS connections and stored at rest on password-protected, encrypted servers on the authors' institutional computing infrastructure.
Access to the raw dataset (including any free-form queries prior to PII scrubbing) was restricted to the listed authors. The compensation file containing Prolific IDs was stored separately from response data, accessible only to the authors. The de-identified dataset---comprising Likert ratings, pairwise preference selections, PII-scrubbed queries, interaction timestamps, and per-session identifiers---is retained for research purposes. Prior to any public release, the dataset will undergo an additional manual review pass to confirm that no inadvertently included identifying information remains.

\subsection{User Interface Screenshot Examples}

\begin{figure}[H]
    \centering
    \begin{subfigure}[t]{0.48\linewidth}
        \centering
        \includegraphics[width=\linewidth]{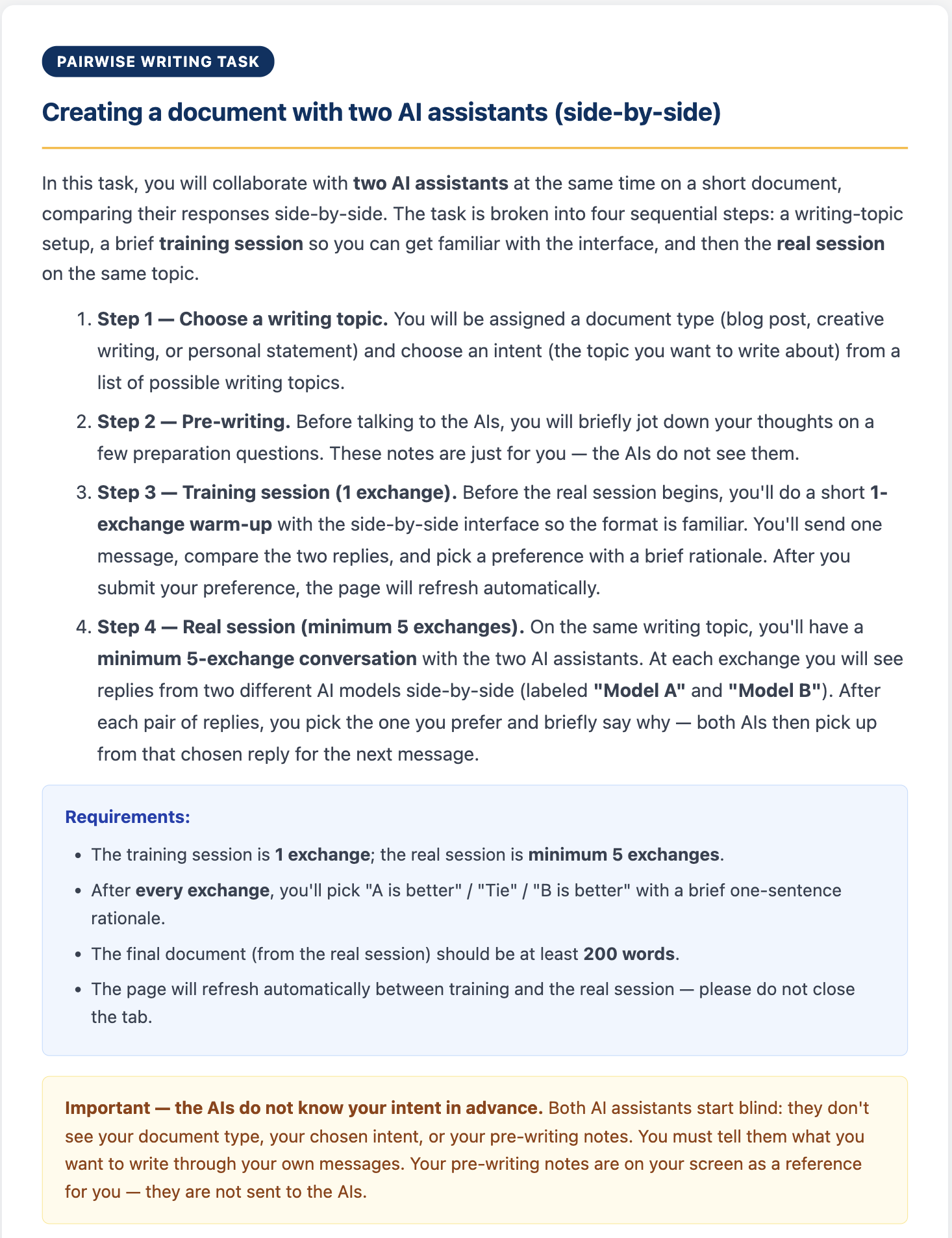}
        \caption{The first instruction page.}
        \label{fig:figure_human_study_step-1-1}
    \end{subfigure}
    \hfill
    \begin{subfigure}[t]{0.48\linewidth}
        \centering
        \includegraphics[width=\linewidth]{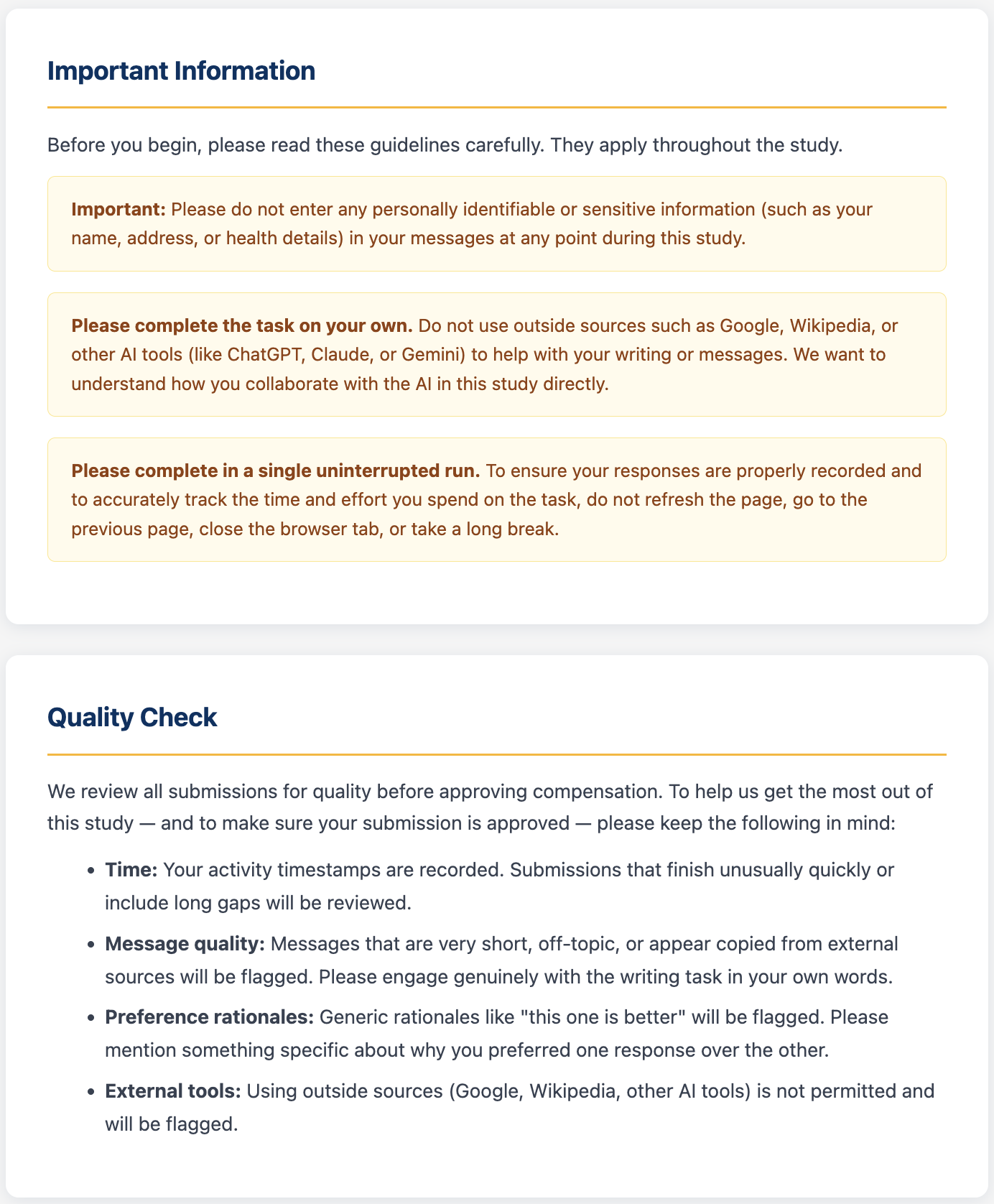}
        \caption{The second instruction page.}
        \label{fig:figure_human_study_step-1-2}
    \end{subfigure}
    \caption{Instruction pages for the human study.}
    \label{fig:figure_human_study_step-1}
    \vspace{\figneg}
\end{figure}

\begin{figure}[H]
    \centering
    \begin{subfigure}[t]{0.48\linewidth}
        \centering
        \includegraphics[width=\linewidth]{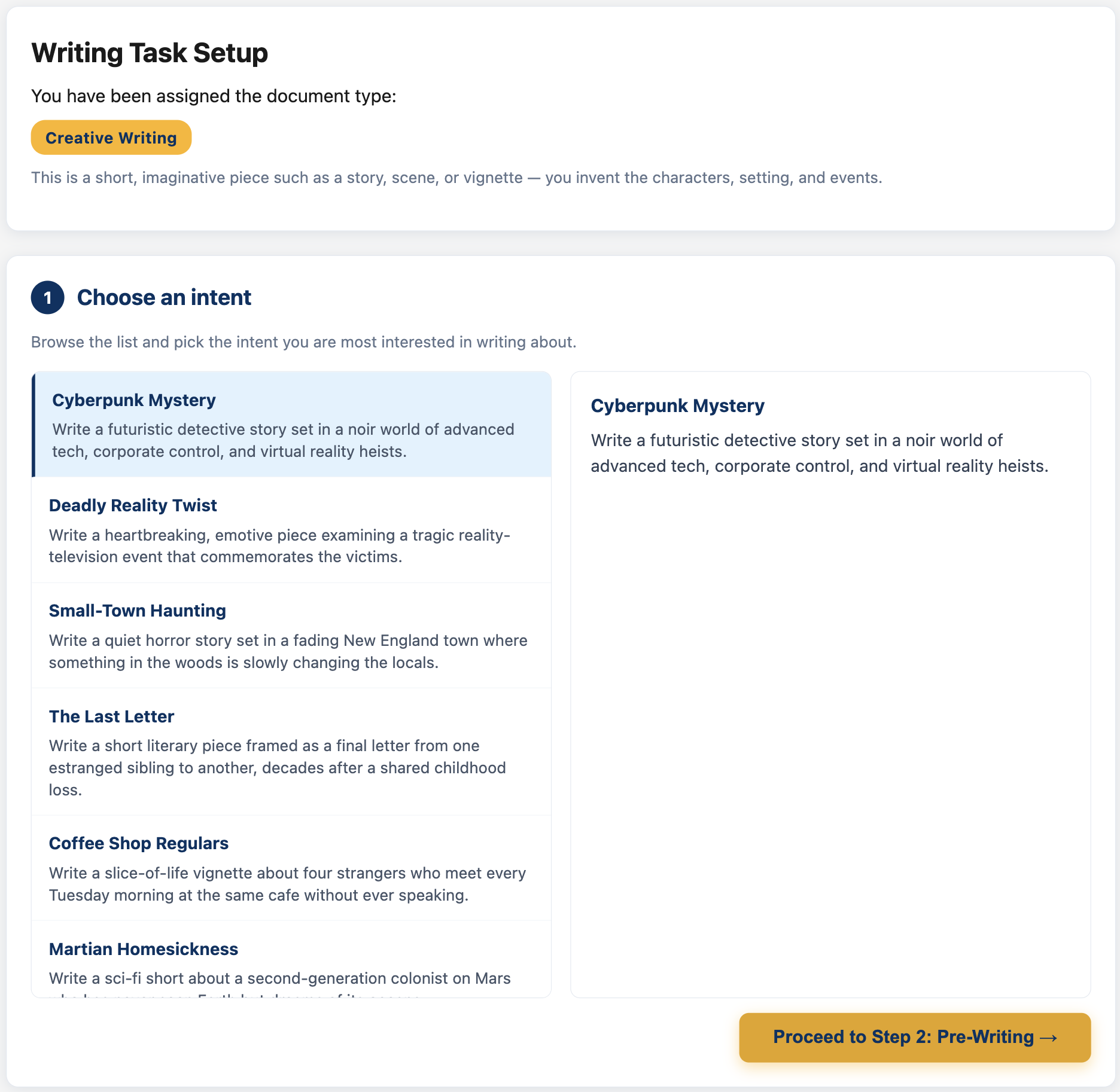}
        \caption{Writing topic selection page.}
        \label{fig:figure_human_study_step-2}
    \end{subfigure}
    \hfill
    \begin{subfigure}[t]{0.48\linewidth}
        \centering
        \includegraphics[width=\linewidth]{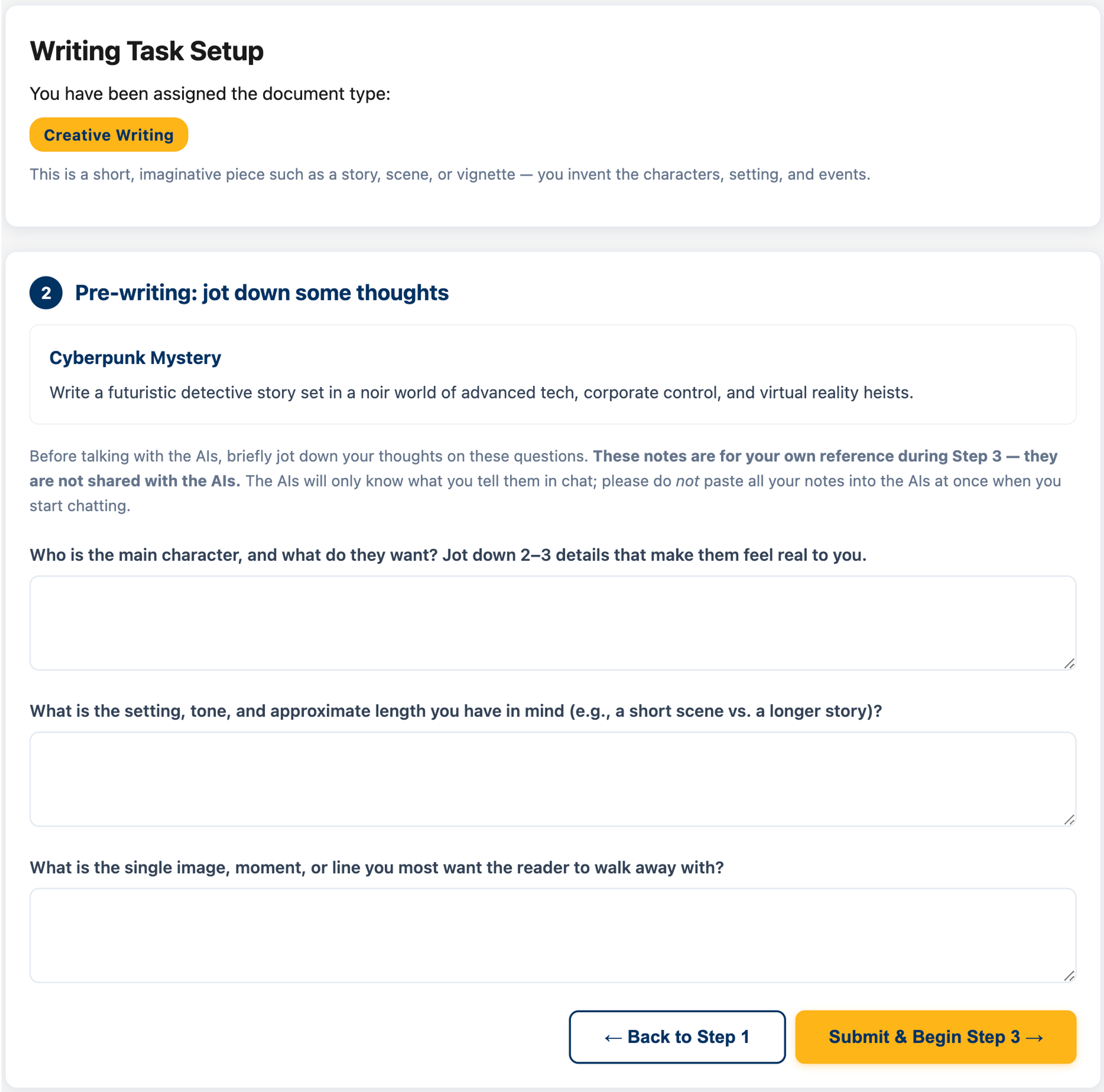}
        \caption{Pre-writing question answering page.}
        \label{fig:figure_human_study_step-3}
    \end{subfigure}
    \caption{Writing topic selection and pre-writing pages.}
    \label{fig:figure_human_study_step-2-3}
    \vspace{\figneg}
\end{figure}

\begin{figure}[H]
    \centering
    \includegraphics[width=1.0\linewidth]{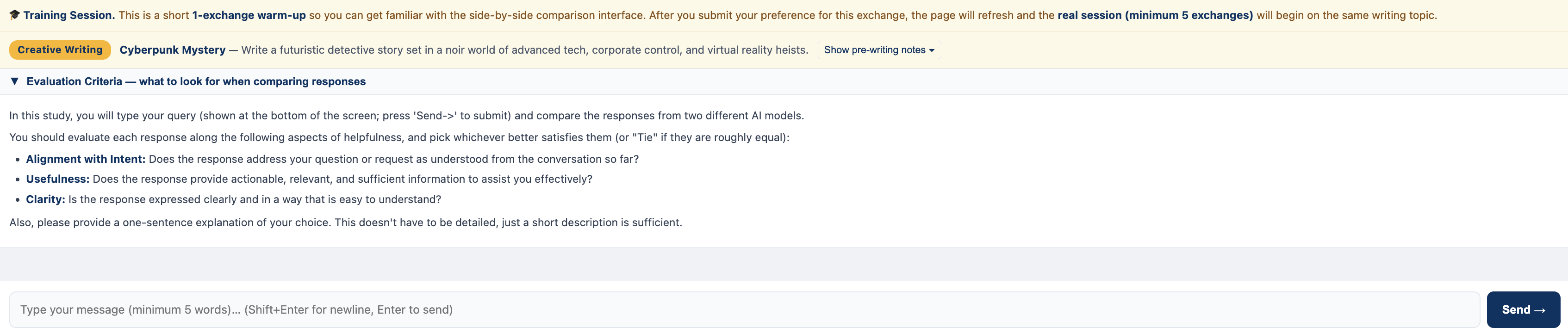}
    \caption{
    Practice session page: participants are asked to try a single-turn conversation to familiarize themselves with the interface.
    \vspace{\figneg}
    }
    \label{fig:figure_human_study_step-4}
\end{figure}

\begin{figure}[H]
    \centering
    \includegraphics[width=0.9\linewidth]{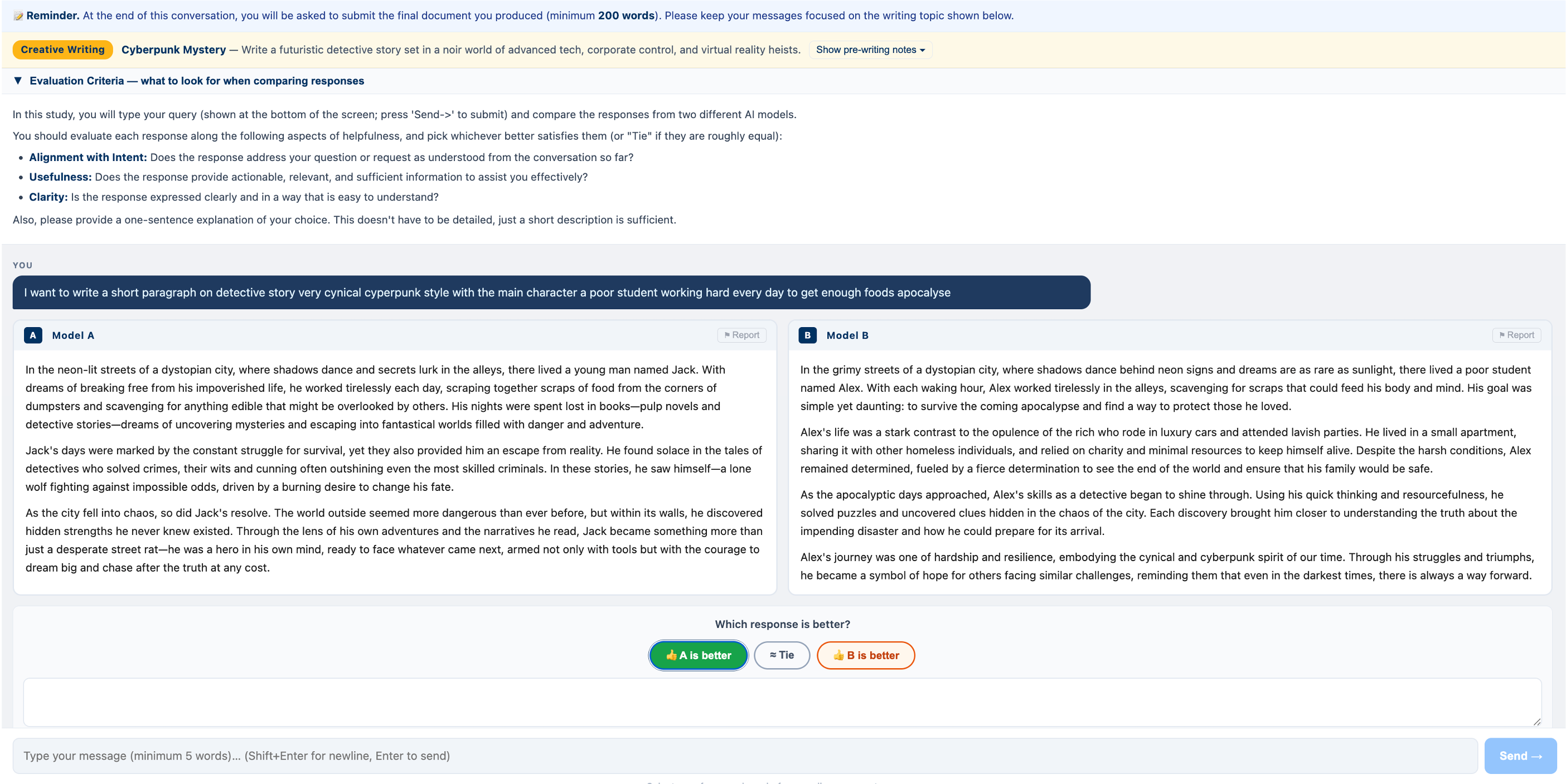}
    \caption{
    Actual conversation session page: participants type in a query, and two anonymized model responses appear side by side. The placement of the two models is randomized at every conversation turn. Once both responses are displayed, participants select their preferred response and provide a brief explanation (at least 12 words).
    }
    \vspace{\figneg}
    \label{fig:figure_human_study_step-5}
\end{figure}

\begin{figure}[H]
    \centering
    \includegraphics[width=1.0\linewidth]{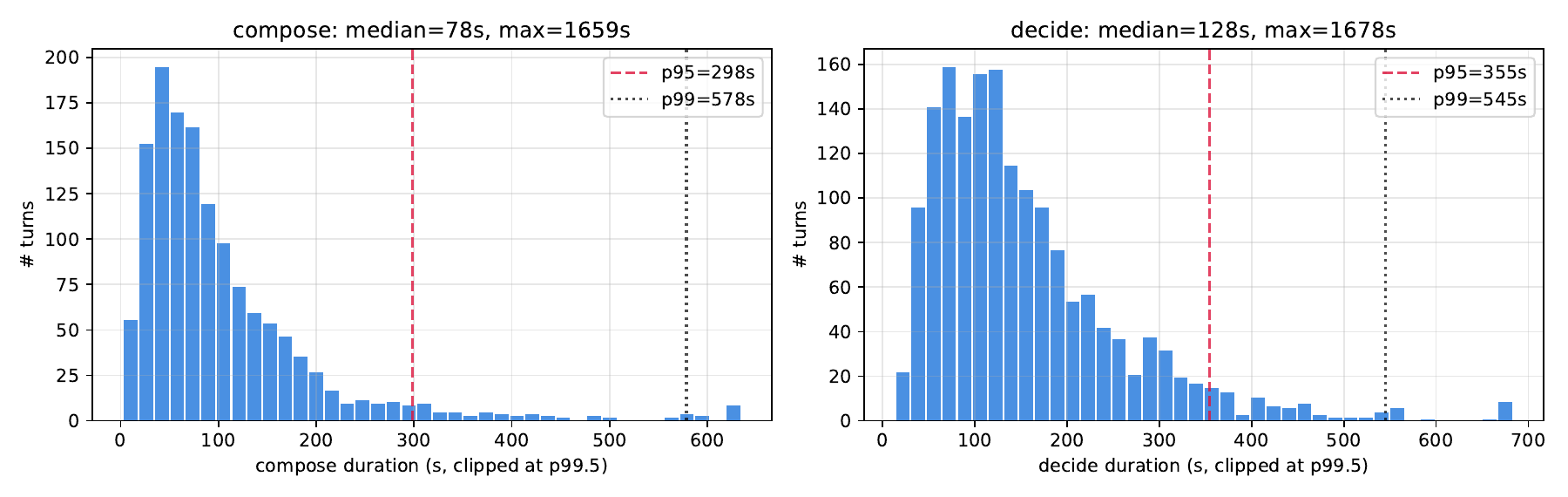}
    \caption{
    Distribution of participant interaction times for composing queries (left) and making pairwise decisions (right).
    }
    \vspace{\figneg}
    \label{fig:figure_human_study_time}
\end{figure}

\subsection{Distribution of Task Completion Times}
\autoref{fig:figure_human_study_time} shows the distribution of times participants took to compose queries (left) and to make pairwise decisions (right). Both are measured from interaction timestamps recorded when participants click `send' to submit a query, when the two assistant responses finish generating and are displayed, and when participants click `decide'. The median times for composing and deciding are 78 and 128 seconds, respectively, yielding roughly 20 minutes total for the combined single-turn practice session and five-turn conversation session.
\section{Additional Results}
\label{appendix:sec:additional_results}
\vspace{\myneg}

\subsection{Comparison With Single-Turn RL}
\label{sec:singleturn}
\vspace{\myneg}

We also consider an assistant training method that does not require a user simulator.
A widely adopted paradigm in RL from Human Feedback (RLHF) is single-turn RL
\citep{christiano2017deep, ouyang2022training, bai2022training, lambert2025reinforcement},
a contextual bandit in which the assistant generates a next-turn response conditioned
on a conversation history, and feedback on that response provides the training signal.
Concretely, given a length-$n$ human--LLM conversation, we construct $n$ training instances:
the $t$-th instance uses $o_t = \{u_1, a_1, \cdots, u_t\}$ as context,
and the assistant generates $a_t$ conditioned on $o_t$.
We use the same LLM judges, scoring rubric, and RL algorithm as in the multi-turn setting (Section~\ref{sec:experiments:setup}).

Single-turn RL has the advantage of conditioning on real human--LLM conversation
prefixes, whereas user simulators retain (at most) the first user turn from real
data and generate the rest.
However, this realism comes with two potential downsides.
First, the conditioning context is generated off-policy
\citep{levine2020offline, gao2024regressing, shani2024multi}: $o_t$ contains assistant responses from the data-collection model (e.g., GPT-3.5) rather
than from the policy being trained.
Second, the policy is optimized for a single-turn reward and is not
incentivized to take actions whose payoff materializes only in later turns, such
as information-gathering actions (asking clarifying questions) \citep{zhang2024modeling}.
Given these trade-offs, one of our secondary research questions asks how the two
training methods compare.
In particular, we investigate whether simulator-based training can surpass
single-turn RL as simulators become more human-like, thereby narrowing the realism
gap user simulators possess.

\newcommand{\posdelta}[1]{\textcolor{ForestGreen}{\small #1}}
\newcommand{\negdelta}[1]{\textcolor{BrickRed}{\small #1}}
\newcommand{\ci}[1]{{\scriptsize$\,{\pm}#1$}}

\begin{table}[t]
    \centering
    \small
    \setlength{\tabcolsep}{2.5pt}
    \renewcommand{\arraystretch}{1.3}
    \caption{
        Comparison of the \nouser-trained assistant with assistants trained with user simulators.
        \textbf{(Left)}
        WildBench win rate of \nouser-trained assistant against others:
        it wins all but the \sftmodel-trained one.
        \textbf{(Right)}
        Cross-simulator eval.
        \sftmodel-trained assistant outperforms \nouser-trained assistant in every test-time simulator;
        \roleplaythree-trained assistant performs comparably or better than \nouser-trained one only when paired with role-playing LLMs.
    }
    \begin{subtable}[t]{0.30\linewidth}
        \centering
        \label{tab:wb_static_orig}
        \begin{tabular}{r| c}
            \toprule
            \textit{Win rate} & \multirow{2}{*}{\nouser} \\
            \textit{Against} [\%] & \\[0.55em]
            \midrule
            Initial        & 67.9 \scriptsize{$\pm$ 2.6} \\
            \roleplayone   & 63.0 \scriptsize{$\pm$ 2.7} \\
            \roleplaytwo   & 57.3 \scriptsize{$\pm$ 2.7} \\
            \roleplaythree & 57.9 \scriptsize{$\pm$ 2.7} \\
            \sftmodel      & 49.3 \scriptsize{$\pm$ 2.8} \\
            \bottomrule
        \end{tabular}
    \end{subtable}
    \hspace{0.5em}
    \begin{subtable}[t]{0.65\linewidth}
        \centering
        \label{tab:cross_eval_static_only}
        \begin{tabular}{r| ccc cc}
            \toprule
            & \multicolumn{3}{c}{\textbf{Assistant trained with}}
            & \multicolumn{2}{c}{$\Delta$ \textbf{vs.\ \nouser}} \\
            \cmidrule(lr){2-4} \cmidrule(lr){5-6}
            \textbf{Eval with}
                & \nouser
                & \roleplaythree
                & \sftmodel
                & \roleplaythree
                & \sftmodel \\
            \midrule
            \roleplayone
                & 87.8\ci{0.2}
                & 87.6\ci{0.2}
                & \textbf{88.7}\ci{0.2}
                & \negdelta{-0.2}
                & \posdelta{+0.9} \\
            \roleplaytwo
                & 85.3\ci{0.2}
                & 86.3\ci{0.2}
                & \textbf{87.5}\ci{0.2}
                & \posdelta{+1.0}
                & \posdelta{+2.2} \\
            \roleplaythree
                & 83.3\ci{0.2}
                & 83.6\ci{0.2}
                & \textbf{84.8}\ci{0.2}
                & \posdelta{+0.3}
                & \posdelta{+1.5} \\
            \sftmodel
                & 80.9\ci{0.3}
                & 80.3\ci{0.3}
                & \textbf{82.5}\ci{0.3}
                & \negdelta{-0.6}
                & \posdelta{+1.6} \\
            \sftmodeltwo
                & 80.9\ci{0.3}
                & 76.8\ci{0.3}
                & \textbf{82.8}\ci{0.2}
                & \negdelta{-4.1}
                & \posdelta{+1.9} \\
            \bottomrule
        \end{tabular}
    \end{subtable}
    \label{tab:static_comparison}
\end{table}

\textbf{Results.}
We denote \nouser the single-turn RL;
we compare the \nouser-trained assistant against assistants trained with user simulators in the multi-turn setting.
\autoref{tab:static_comparison} left reports pairwise win rate of the \nouser-trained assistant against every other assistant on WildBench-v2.
\nouser wins against all role-playing variants and the initial assistant (57.3--67.9\%),
but is comparable to the \sftmodel-trained assistant (49.3 $\pm$ 2.8\%);
checklist item satisfaction rate shows the same ordering, with \nouser (61.0\%) outperforming every other assistant except the one trained with \sftmodel (61.5\%).
\autoref{tab:static_comparison} (right) shows the cross-user evaluation of
\nouser, \roleplaythree, and \sftmodel.
The \sftmodel-trained assistant outperforms the \nouser-trained assistant under
every test-time user simulator (all $\Delta > 0$).
We posit that, although \nouser is always conditioned on real human--LLM
conversation history, its single-turn objective does not encourage policies that
improve trajectory-level outcomes.
By contrast, the \roleplaythree-trained assistant---the strongest among those
trained with role-playing LLM simulators---underperforms \nouser.
Taken together, these results suggest that multi-turn training alone is not
sufficient to surpass single-turn RL; only training with a
sufficiently human-like user simulator (\sftmodel) consistently matches or improves
over single-turn training.

\textbf{Note.}
The WildBench evaluation structurally favors \nouser:
its training input distribution (static offline human--LLM conversation prefixes) exactly matches the WildBench protocol,
in which an assistant generates a single-turn response given a fixed context.
Moreover, because each assistant produces only the final response, WildBench cannot reward capabilities that emerge from shaping the conversation trajectory through earlier actions (e.g., asking clarifying questions).
Despite this distributional advantage for \nouser, the \sftmodel-trained assistant performs comparably or better,
suggesting that exposure to realistic human utterance patterns during training can improve single-turn response quality even under input distribution shift.

\subsection{User Simulator Fidelity Metrics and Trained Assistant Performance}
\label{appendix:sec:usersim_fidelity}

\begin{figure}[t]
    \centering
    \includegraphics[width=1.00\linewidth]{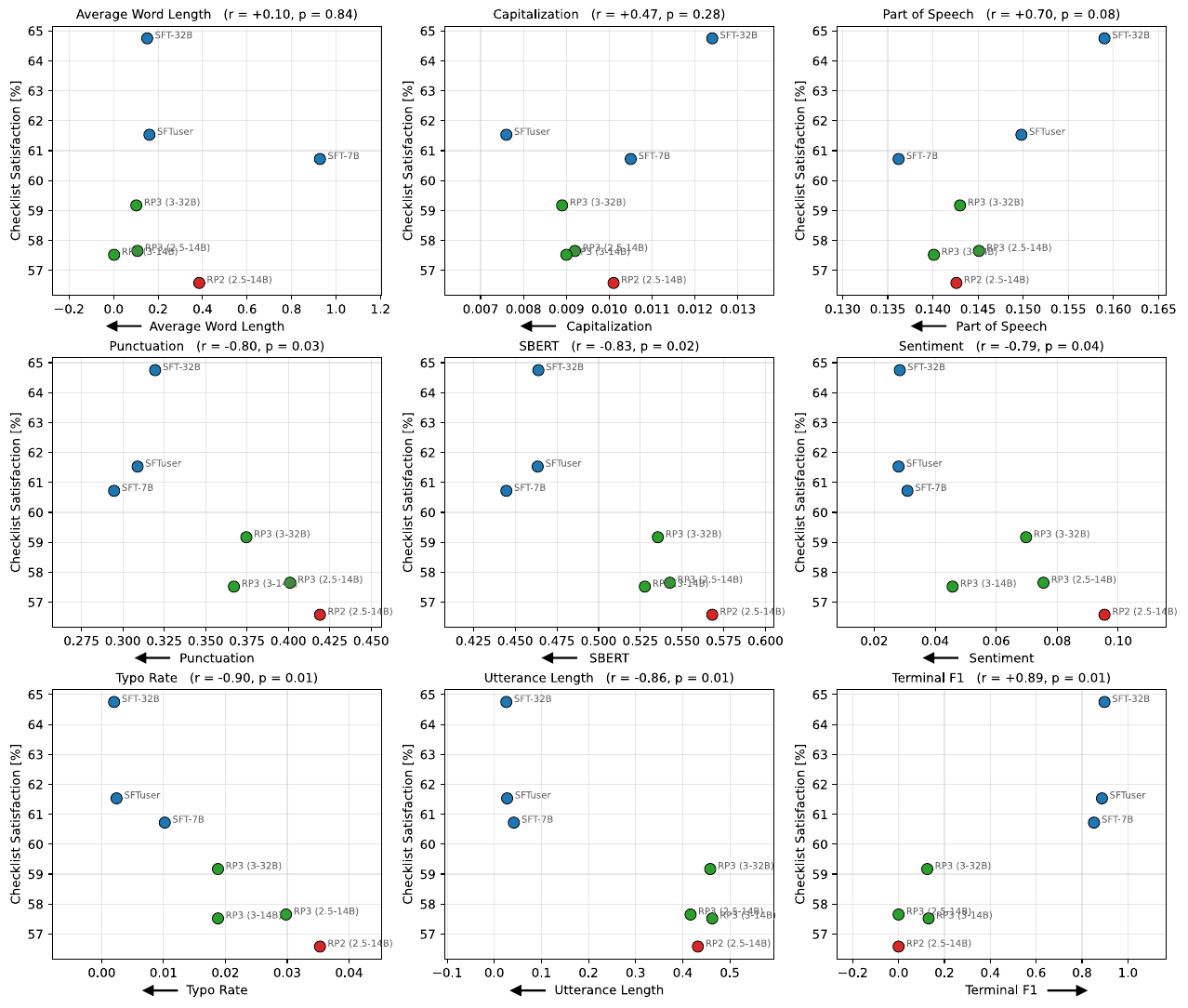}
    \caption{
    Per-metric comparison of the user simulators from Section~\ref{sec:experiments}.
    Each panel reports a fidelity metric adapted from \citet{ivey2024robotic}, computed by replaying WildBench-v2 conversations turn-by-turn and comparing each simulator-generated utterance to the ground-truth human utterance.
    Arrows indicate the direction of improvement.
    The y-axis shows the WildBench checklist satisfaction rate of the assistant trained against each simulator (\autoref{tab:wildbench_pairwise}).
    }
    \vspace{\figneg}
    \label{fig:usersim_metric_vs_wildbench}
\end{figure}

We adapt the simulator fidelity metrics of \citet{ivey2024robotic} to the simulators used in Section~\ref{sec:experiments},
and relate them to the WildBench checklist satisfaction rate of assistants trained with each simulator
(\autoref{tab:wildbench_pairwise} and \ref{tab:wildbench_scaling_combined}).
\autoref{fig:usersim_metric_vs_wildbench} reports nine metrics spanning lexical, syntactic, semantic, stylistic, and conversation-control properties of simulated user utterances:
\vspace{\myneg}
\begin{itemize}[leftmargin=2em, itemsep=2pt, parsep=0pt]
    \item \textbf{Average word length:}
    mean characters per word, compared to the ground-truth human utterance via $\ell_1$ distance. Captures whether the simulator favors shorter, more colloquial words or longer, more formal ones. Values closer to 0 indicate a closer match to human's average word length.
    \item \textbf{Utterance length:}
    Jensen--Shannon divergence (JSD) between the simulator's and the human's word-count distributions. Values closer to 0 indicate more similar utterance-length distributions.
    \item \textbf{Typographical error rate:}
    fraction of tokens flagged as misspellings by \texttt{pyspellchecker}, compared to human users via $\ell_1$ distance. Real WildChat users produce nontrivial typo rates; values closer to 0 indicate the simulator typos at a similar rate.
    \item \textbf{Part-of-speech (POS) distribution:}
    JSD between the empirical POS-tag distributions (from \texttt{spaCy}) of the simulator and human utterances. Lower is closer to human.
    \item \textbf{Punctuation distribution:}
    JSD over the distribution of punctuation characters. Differentiates simulators that over- or under-use markdown-like structure (bullets, em-dashes, etc.).
    \item \textbf{Capitalization:}
    JSD over the distribution of capitalization patterns. Real users frequently write in all-lowercase or mixed case; values closer to 0 indicate the simulator matches the human distribution of capitalization patterns.
    \item \textbf{Sentiment distribution:}
    JSD between the simulator's and human's positive/neutral/negative sentiment distributions, classified by \texttt{distilbert-base-multilingual-cased-sentiments-student}.
    \item \textbf{SBERT semantic distance:}
    $1 - \cos$ similarity between the user simulator's and human's sentence embeddings from \texttt{all-MiniLM-L6-v2} \citep{reimers2019sentence}. Captures whether the simulator's utterance is semantically close to what the human actually said next.
    Lower is closer to human.
    \item \textbf{Terminal F1:}
    binary F1 over continue/terminate decisions. At each user turn the simulator either emits an utterance (continue) or an end-conversation special token (terminate); ground truth is ``continue'' for every turn within the human conversation and ``terminate'' once it has ended. Higher is better.
\end{itemize}

\textbf{Measurement.}
We use WildBench-v2 conversations \citep{lin2024wildbench} for the reference human utterance distribution.
For each length-$n$ human--LLM conversation $\{u_1, a_1, \ldots, u_n, a_n\}$, we replay turn-by-turn:
at each user position $t \in \{2, \ldots, n\}$ we condition user simulator on the prefix $\{u_1, a_1, \ldots, u_{t-1}, a_{t-1}\}$ and elicit a candidate utterance $\hat{u}_t$, then compare $\hat{u}_t$ to the ground-truth $u_t$ on the metrics above.
For terminal F1, we additionally elicit one more decision past the final human turn ($t = n+1$), where the ground-truth label is ``terminate.''
Each metric is averaged across all turns and all conversations.

\textbf{Results.}
The learned user simulator (\sftmodel) more closely matches human utterances than the role-playing simulators across multiple metrics:
punctuation and sentiment distributions, SBERT semantic distance, typo rate, utterance-length distribution, and terminal F1.
The gap is largest on terminal F1:
role-playing simulators almost never emit the end-conversation token (consistent with \citep{ivey2024robotic}),
whereas \sftmodel correctly predicts termination most of the time (consistent with \citep{naous2025flipping}).
We note that on three remaining metrics (average word length, capitalization, and POS distribution) the results are inconclusive, with no clear advantage for either types of user simulators.

\textbf{Relating fidelity to downstream performance.}
One can define many plausible fidelity metrics in utterance space, and it is not
obvious a priori which would predict downstream assistant quality.
Comparing \autoref{fig:usersim_metric_vs_wildbench} against the WildBench
results (\autoref{tab:wildbench_pairwise} and \ref{tab:wildbench_scaling_combined}),
we find that the relationship is uneven: the ordering of simulators by some metrics (e.g., SBERT semantic distance, distribution of the utterance length, terminal F1) broadly aligns with their ordering
by checklist satisfaction, while others (e.g., average word length, capitalization) are
not correlated to downstream performance. We read this as a
cautionary observation rather than a validation of utterance-level fidelity:
the space of such metrics is large, but only a subset
appears informative about the quality of the assistant ultimately trained
against the simulator. This argues against treating any single utterance-level
metric as a proxy for simulator usefulness, and motivates the direct
end-to-end evaluations reported in Section~\ref{sec:experiments:wildbench}
as the more reliable signal.

\subsection{Result with Qwen2.5-1.5B-Instruct}

Here we repeat the experiments in Section~\ref{sec:experiments:wildbench},
starting from Qwen2.5-1.5B-Instruct as the initial LLM assistant checkpoint.
\autoref{tab:wildbench_pairwise_small} presents the WildBench pairwise result mirroring \autoref{tab:wildbench_pairwise},
showing the same trend as in the Qwen2.5-3B-Instruct case where more human-like user simulators yield stronger trained assistants on WildBench.

\begin{table}[t]
    \centering
    \small
    \setlength{\tabcolsep}{2.0pt}
    \renewcommand{\arraystretch}{1.3}
    \caption{
        Tie-accounted pairwise win rate matrix and checklist satisfaction rate on WildBench-v2.
        The entry in the $i$-th row, $j$-th column indicates the win rate of
        the assistant trained with $j$ against the one trained with $i$;
        values after $\pm$ denote the half-width of the 95\% Wald confidence interval.
    }
    \begin{tabular}{r *{5}{w{c}{4.5em}}}
        \toprule
        \textit{Win} [\%]
        & \rot{Initial} & \rot{\roleplayone} & \rot{\roleplaytwo} & \rot{\roleplaythree} & \rot{\sftmodel} \\
        \midrule
        Initial
            & \diagcell
            & \pcell{54.9}{2.8}
            & \pcell{57.9}{2.7}
            & \pcell{60.2}{2.7}
            & \pcellbf{67.0}{2.6} \\
        \roleplayone
            & \pcell{45.1}{2.8}
            & \diagcell
            & \pcell{53.1}{2.8}
            & \pcell{55.6}{2.8}
            & \pcellbf{62.2}{2.7} \\
        \roleplaytwo
            & \pcell{42.1}{2.7}
            & \pcell{46.9}{2.8}
            & \diagcell
            & \pcell{52.2}{2.8}
            & \pcellbf{59.3}{2.7} \\
        \roleplaythree
            & \pcell{39.8}{2.7}
            & \pcell{44.4}{2.8}
            & \pcell{47.8}{2.8}
            & \diagcell
            & \pcellbf{56.8}{2.7} \\
        \sftmodel
            & \pcell{33.0}{2.6}
            & \pcell{37.8}{2.7}
            & \pcell{40.7}{2.7}
            & \pcell{43.2}{2.7}
            & \diagcell \\
        \midrule
        \textit{Satisfy} [\%]
            & 33.5\scriptsize{$\pm$0.9}
            & 36.1\scriptsize{$\pm$0.9}
            & 37.8\scriptsize{$\pm$0.9}
            & 39.1\scriptsize{$\pm$0.9}
            & \textbf{42.9}\scriptsize{$\pm$1.0} \\
        \bottomrule
    \end{tabular}
    \vspace{\figneg}
    \label{tab:wildbench_pairwise_small}
\end{table}

\end{document}